\documentclass[lettersize,journal]{IEEEtran}
\usepackage{amsmath,amsfonts}
\usepackage{algorithmic}
\usepackage{algorithm}
\usepackage{array}
\usepackage{textcomp}
\usepackage{stfloats}
\usepackage{url}
\usepackage{verbatim}
\usepackage{graphicx}
\usepackage{cite}
\usepackage{eqnarray,amsmath}
\usepackage{amssymb}
\usepackage{tabularx}
\usepackage{adjustbox}
\usepackage{amssymb}
\usepackage{pifont}
\usepackage{float}

\usepackage{amsmath}
\usepackage{amssymb}
\usepackage[table]{xcolor}
\usepackage[pagebackref=true,breaklinks=true,colorlinks=true,bookmarks=false,linkcolor={red!50!black},urlcolor={darkgray!80!black},citecolor={green!50!black}]{hyperref}
\usepackage[group-separator={,}]{siunitx}
\usepackage{multirow}
\usepackage{tabularx}
\usepackage{booktabs} 
\usepackage{makecell}
\usepackage{graphbox}
\usepackage{adjustbox}
\usepackage{footmisc}
\usepackage{bm}
\usepackage{caption}
\usepackage{arydshln}
\usepackage{dashrule}
\usepackage{subcaption}
\usepackage{enumitem}
\setlist[itemize]{noitemsep,nolistsep}

\usepackage[capitalize]{cleveref}
\crefname{section}{Sec.}{Secs.}
\Crefname{section}{Section}{Sections}
\Crefname{table}{Table}{Tables}
\crefname{table}{Tab.}{Tabs.}
\Crefname{figure}{Figure}{Figures}
\crefname{figure}{Fig.}{Figs.}
\Crefname{equation}{Equation}{Equations}
\crefname{equation}{Eq.}{Eqs.}

\usepackage{xspace}
\makeatletter
\DeclareRobustCommand\onedot{\futurelet\@let@token\@onedot}
\def\@onedot{\ifx\@let@token.\else.\null\fi\xspace}

\def\etal{\emph{et al}\onedot}
\makeatother

\usepackage{dsfont}

\colorlet{lightpink}{pink!35}
\colorlet{lightcyan}{cyan!20}
\colorlet{red}{red!80}
\colorlet{blue}{blue!80}
\colorlet{green}{green!60!black}

\newcolumntype{C}[1]{>{\centering\arraybackslash}p{#1}}
\newcolumntype{L}[1]{>{\raggedleft\arraybackslash}p{#1}}
\newcolumntype{R}[1]{>{\raggedright\arraybackslash}p{#1}}

\newcommand{\RNum}[1]{\uppercase\expandafter{\romannumeral #1\relax}}

\graphicspath{{images/}}
\usepackage[switch]{lineno}

\begin{document}

\title{Diffusion-based Facial Aesthetics Enhancement with 3D Structure Guidance}

\author{
Lisha Li,
Jingwen Hou,
Weide Liu,
Yuming Fang,~\IEEEmembership{Senior Member,~IEEE},
Jiebin Yan
\thanks{This work was supported in part by the National Key Research and Development Program of China under Grant 2023YFE0210700; in part by the National Natural Science Foundation of China under Grant 62132006, Grant 62461028, Grant 62441203, Grant 62162029, and Grant 62311530101; in part by the Natural Science Foundation of Jiangxi Province of China under Grant 20223AE191002, Grant 20232BAB202001 and Grant 20243BCE51139; and in part by the project funded by China Postdoctoral Science Foundation under Grant 2024T170364; and in part by Jiangxi Provincial Key Laboratory of Multimedia Intelligent Processing under Grant 2024SSY03141. \textit{(Corresponding author: Jingwen Hou.)}}
\thanks{Lisha Li, Jingwen Hou, Yuming Fang, and Jiebin Yan are with the School of Computing and Artificial Intelligence, Jiangxi University of Finance and Economics, China; Weide Liu is with the Boston Children’s Hospital and Harvard Medical School, Boston, MA. 
(e-mail: lilisha19961030@163.com; jingwen003@e.ntu.edu.sg; weide001@e.ntu.edu.sg; fa0001ng@e.ntu.edu.sg; jiebinyan@foxmail.com)}
}

\markboth{IEEE Transactions on Image Processing,~Vol.X, No.X, Year}%
{Shell \MakeLowercase{\etal}: A Sample Article Using IEEEtran.cls for IEEE Journals}


\maketitle

\begin{abstract}
Facial Aesthetics Enhancement (FAE) aims to improve facial attractiveness by adjusting the structure and appearance of a facial image while preserving its identity as much as possible. Most existing methods adopted deep feature-based or score-based guidance for generation models to conduct FAE. Although these methods achieved promising results, they potentially produced excessively beautified results with lower identity consistency or insufficiently improved facial attractiveness. To enhance facial aesthetics with less loss of identity, we propose the Nearest Neighbor Structure Guidance based on Diffusion (NNSG-Diffusion), a diffusion-based FAE method that beautifies a 2D facial image with 3D structure guidance. Specifically, we propose to extract FAE guidance from a nearest neighbor reference face. To allow for less change of facial structures in the FAE process, a 3D face model is recovered by referring to both the matched 2D reference face and the 2D input face, so that the depth and contour guidance can be extracted from the 3D face model. Then the depth and contour clues can provide effective guidance to Stable Diffusion with ControlNet for FAE. Extensive experiments demonstrate that our method is superior to previous relevant methods in enhancing facial aesthetics while preserving facial identity.
\end{abstract}

\begin{IEEEkeywords}
Facial Aesthetics Enhancement, Stable Diffusion.
\end{IEEEkeywords}

\section{Introduction}
\label{sec:intro}
\IEEEPARstart{F}{a}cial Aesthetics Enhancement (FAE) (or facial beautification) aims to adjust the facial structure and appearance of a facial image to improve its aesthetics while preserving its identity.
In recent years, research indicated that facial attractiveness plays a significant role in enabling individuals to obtain more social opportunities \cite{hedman2022effect,lebedeva2021mebeauty}. Therefore, beauty cameras have gradually become widely used in applications of human-computer interaction, entertainment, live streaming, e-commerce, and popular social media platforms such as TikTok and Instagram. 

Recently, promising results have been achieved by deep learning-based FAE methods, including GAN-based methods \cite{xia2021open,karras2020training, diamant2019beholder} or diffusion-based methods \cite{huang2023collaborative,liu2022compositional,wang2024instantid,li2023photomaker}. Typically, guidance is required when FAE is conducted, so that the facial aesthetics can be eventually improved.
Forms of guidance adopted by previous deep FAE methods includes: 1) deep visual or textual feature-based guidance \cite{zhou2020gan,karras2020training, xia2021open,liu2022compositional,huang2023collaborative,valevski2023face0,ye2023ipadapter,li2023photomaker}; 2) score-based guidance \cite{diamant2019beholder,liu2019understanding}; 3) explicit face features such as facial landmarks \cite{chen2023aep,wang2024instantid}. 
However, these FAE methods may encounter several limitations, leading to loss of original identity or insufficiently beautified results: 1) Reliance only on deep features for guidance potentially leads to excessively beautified results due to the high degree of freedom in introducing generative details. 2) Using only 2D facial features for guidance may not adequately capture structural details such as outer and inner contours, proportions of facial parts, highlights, and shadows depicting facial structures. 3) Misalignment between the pose of input and reference face can lead to incorrect or ineffective guidance, or even mislead the generation process, resulting in faces with inconsistent poses. 4) Manual processes, such as prompt engineering \cite{radford2021learning}, are often required to provide representations as references by these previous methods.

\begin{figure}
    \centering
    \includegraphics[width=1\linewidth]{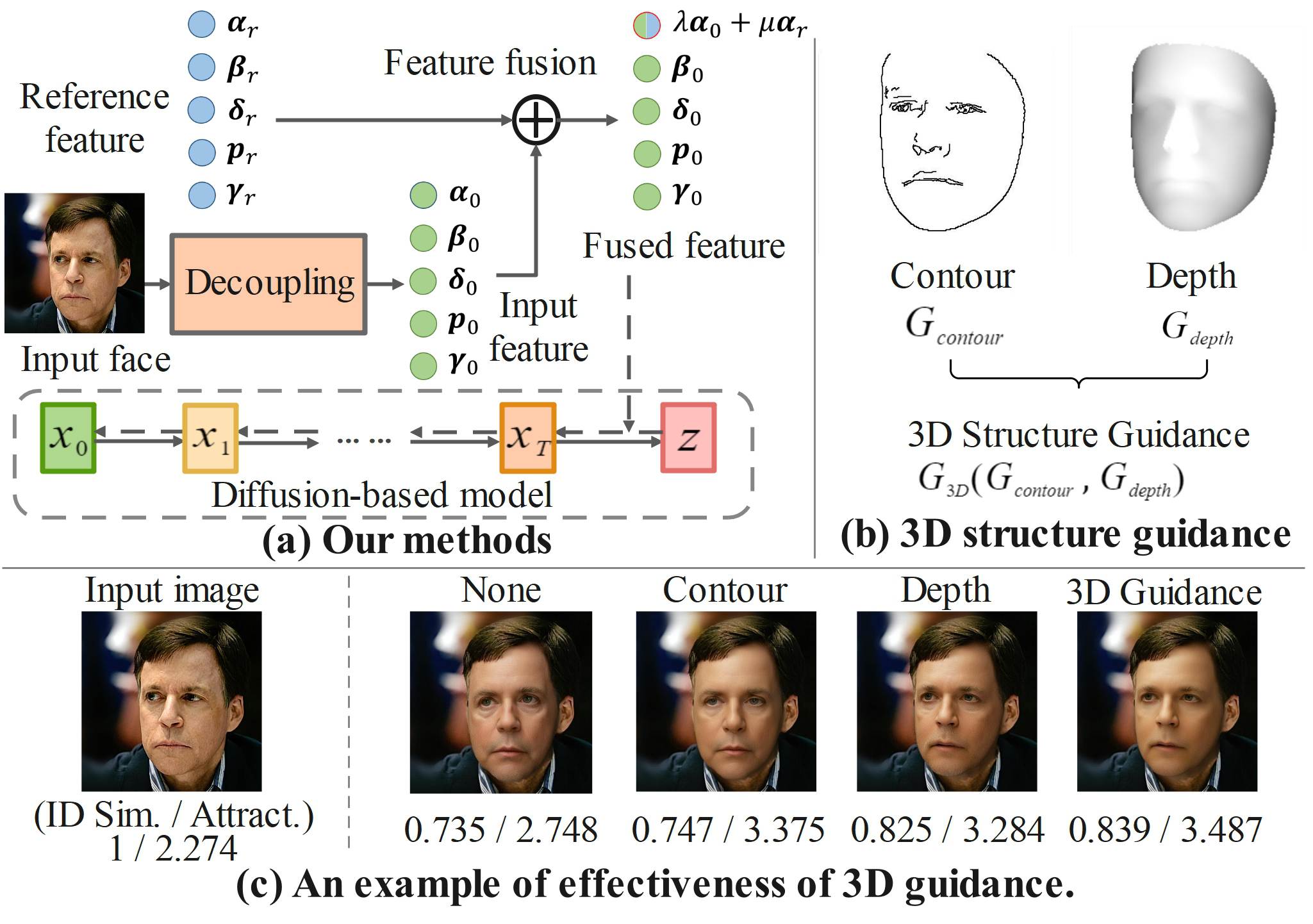}
    \caption{
    Our method provides effective beautification guidance by: 
    (a) Recovering 3DMM model \cite{blanz2023morphable} from the 2D reference image and 2D reference image, and combining them to construct a 3D prototype;
    (b) Extracting contour guidance and depth guidance from the 3D prototype.
    Our method improves facial attractiveness with the highest ID similarity (compared to the input) when combining the contour and the depth guidance to Stable Diffusion \cite{rombach2022high} with ControlNet \cite{zhang2023adding}, as shown in (c).}
    \label{fig:teaser}
    \vspace{-6mm}
\end{figure}

Our method is designed to provide structural guidance besides appearance guidance that can be given by features directly extracted from 2D images \cite{ye2023ipadapter}. 
To effectively improve facial beauty and maintain identity consistency, we introduce Nearest Neighbor Structure Guidance based on Diffusion Model (NNSG-Diffusion). The proposed method consists of three modules: a Nearest Neighbor Face Searching (NNFS) module, a Facial Guidance Extraction (FGE) module, and a Face Beautification (FB) module.
The NNFS module first finds a reference face nearest to the input face in structure within our aesthetic prototype database, which consists of portraits with high aesthetic scores.
Upon finding the reference face, the FGE module extracts the guidance for facial contour and depth clues from the 3D face model reconstructed from the reference face image and the input face image.
Finally, the FB module utilizes Stable Diffusion (SD) \cite{rombach2022high} with ControlNet \cite{zhang2023adding} to beautify the input face based on 3D structure guidance consisting of the depth and the contour guidance. 
To focus on improving facial structure, the NNFS module uses decoupled identity feature to search for the most similar reference face.
Among all possible reference faces in our database, using the closest face as guidance allows the downstream FAE process to introduce fewer changes to the input's identity but ensures improvements in facial aesthetics.

As presented in Fig. \ref{fig:teaser}(a), our method blends the parameterized face model of the referenced face with that of the input face to obtain 3D structure guidance (as shown in Fig. \ref{fig:teaser}(b)) for the later FAE generation process. 
Fig. \ref{fig:teaser}(c) presents a comparison among the results with or without 3D structure guidance. 
On the one hand, depth guidance can provide the SD model with information about facial spatial structure and pose to depict the 3D structure of the enhanced face. On the other hand, the contour guidance can provide the diffusion model with information about the contour of the inner and outer parts of the face, inducing results with similar facial part proportions and outlines as the reference face. These two forms of guidance are then combined to provide the SD model as 3D structure guidance.

To summarize, the contributions of our work can be summarized as follows:

\begin{itemize}
\item{We propose a new FAE method called NNSG-Diffusion, which accurately controls FAE process with SD using the proposed 3D structure guidance. This effectively improves facial attractiveness while relieving the problem of excessive facial beautification and loss of identity.}
\item {We propose the NNFS module to search for the reference face that is closest to the input face's 3D structure in the aesthetic prototype database, helping to preserve facial identity information from the perspective of FAE guidance. Furthermore, we establish an aesthetic prototype database derived from the CelebA database \cite{liu2015faceattributes} to support the proposed face searching.}
\item{We propose the FGE module for 3D structure guidance extraction. The module helps to effectively introduce beautification guidance from the reference face with less turbulence on identity information and mitigates undesired guidance caused by inconsistency between the poses or expressions of the input and the reference.}
\end{itemize}

\section{Related Works}

\subsection{Facial Aesthetics Enhancement}

FAE aims to adjust facial structure and appearance of a facial image to improve its aesthetics while preserving its identity.
Early researchers followed established aesthetic theory to directly manipulate the structure of the face \cite{perrett1994facial}. These theories include concepts like the ``three courtyards and five eyes'', the golden ratio \cite{liang2019golden}, and a symmetrical face. 
These theories believe that human faces obeying these established rules are considered beautiful. 
However, these established rules cannot sufficiently explain the aesthetics of individual variations in facial features and preferences. 

Nowadays, data-driven FAE approaches \cite{leyvand2008data, sun2011face, hu2021facial,zhou2020gan,karras2020training, xia2021open,liu2022compositional,huang2023collaborative,valevski2023face0,ye2023ipadapter,li2023photomaker,diamant2019beholder,chen2023aep,wang2024instantid,liu2019understanding} gradually became popular, enabling more diverse facial features and attributes to be considered. Specifically, FAE methods based on deep learning achieved promising results in recent works \cite{hu2021facial,zhou2020gan,karras2020training, xia2021open,liu2022compositional,huang2023collaborative,valevski2023face0,ye2023ipadapter,li2023photomaker,diamant2019beholder,chen2023aep,wang2024instantid,liu2019understanding}.
These deep FAE methods typically adopted one or more forms of reference information to guide the FAE process, including: 1) visual or textual deep feature-based guidance \cite{zhou2020gan,karras2020training, xia2021open,liu2022compositional,huang2023collaborative,valevski2023face0,ye2023ipadapter,li2023photomaker,liu2019understanding}; 2) score-based guidance \cite{diamant2019beholder,liu2019understanding}; 3) explicit face features such as facial landmarks \cite{chen2023aep,wang2024instantid}. 

Visual or textual features can be used for FAE by adopting GAN or diffusion models. GAN-based methods such as StyleGAN \cite{karras2020training} can be easily adapted to FAE by mixing the features of input and reference face before decoding. 
Other GAN-based methods also attempted to learn a latent distribution of visual feature for beautiful faces \cite{zhou2020gan}, or adopted textual features as conditions for generation \cite{xia2021open}. 
State-of-the-art methods with visual or textual deep feature guidance are mostly based on diffusion model \cite{song2020score,rombach2022high}, which generates images by iteratively refining ``denoised'' feature within a latent representation space for decoding into complete images. 
In the case of FAE, diffusion-based methods \cite{liu2022compositional, huang2023collaborative} can easily adopt deep textual or visual features from a given prompt indicating the direction of facial beautification. 
For example, Composable Diffusion \cite{liu2022compositional} edited images by a set of text-guided diffusion models, each of which was responsible for a part of the image.  
Collaborative Diffusion \cite{huang2023collaborative} adopted text encoding and segmentation masks to collaboratively achieve multi-modal facial generation and editing. These models demonstrate remarkable performance in facial attribute editing. 

Score-based guidance can directly use scores as conditions for generation models, or use scores as clues for filtering relevant representations as generation guidance.
For example, Beholder-GAN \cite{diamant2019beholder} incorporates aesthetic scores as conditions to generate realistic facial images, which is a variant of the Progressive Growing of GANs (PGGAN) \cite{karras2017progressive}.  
Liu \etal \cite{liu2019understanding} studied facial attributes relevant to facial aesthetics according to aesthetic scores, and then attempted to adjust facial aesthetics by adjusting relevant facial attributes with a GAN model.  Besides visual or textual deep feature-based guidance and score-based methods, some methods \cite{chen2023aep,wang2024instantid} also introduced facial landmarks as explicit constraints to control the generated faces. 

However, the above-mentioned methods for FAE have the following drawbacks, leading to loss of original identity or insufficiently beautified results: 1) When only deep features are used as guidance, the model often produces results with loss of identity information due to the high degree of freedom in introducing generative details.
2) Although explicit facial features such as landmarks can also be used, most of these methods only adopt 2D features as guidance and may not sufficiently provide information on generating structural details such as contours of outer and inner contours, proportions of facial parts, facial pose and spatial structure (highlights and shadows); 3) When the reference face is of a different pose from the input face, the reference face may not provide correct guidance to the generation model and even mislead the generation process to generate faces with poses inconsistent with the input ones; 4) For text feature-based guidance, a manual prompt engineering process \cite{radford2021learning} is usually required to achieve satisfying results.

Based on SD \cite{rombach2022high}, the proposed method is designed to relieve the above-mentioned issues in previous FAE methods. First, the proposed method adopts 3D structure guidance of a nearest neighbor aesthetic prototype, allowing for fewer change to the input's facial structure. Second, the guidance of the proposed method is obtained from reconstructed 3D face models, resulting in effective guidance in enhancing the structures of the input face. Third, our guidance is extracted from the reference face model projected onto the same direction as the input face, avoiding misleading information provided by the guidance with inconsistent face poses. Lastly, instead of using manually selected guidance, we propose the NNFS process to automatically find the most appropriate reference face for producing the 3D guidance.

\subsection{ID-Preserving Image Generation}
Recent works \cite{ye2023ipadapter,wang2024instantid,valevski2023face0,hu2021lora} related to identity preservation mainly focus on preserving internal facial appearances (e.g., color and shape of eyes, shape of nose, skin tone) with implicit features as guidance. 
Low-Rank Adaptation (LoRA) \cite{hu2021lora}, a lightweight finetuning approach, incorporated low-rank matrices into the pre-trained diffusion model, so that LoRA can generate images in the identity of the provided finetuning dataset (e.g., images of a specific person). 
PhotoMaker \cite{li2023photomaker} proposed an effective fine-tuning strategy with stacked ID embedding to integrate a few identity images for the preservation of the identity of generated images. 
To avoid further fine-tuning in the identity-oriented generation process,  Face0 \cite{valevski2023face0} modified the face embedding vectors and projected the face embedding to Stable Diffusion’s CLIP space \cite{radford2021learning}.

IP-Adapter \cite{ye2023ipadapter} introduced a plug-in allowing an image to be used as guidance for generation. Specifically, an additional cross-attention module for image embeddings was adopted to avoid the missing image details caused by the merging image and text features. As a result, the facial identity features can be effectively captured by the diffusion model with the IP-Adapter.
Unlike IP-Adapter, FaceStudio \cite{yan2023facestudio} used a prior model \cite{ramesh2022hierarchical} to translate textual descriptions into image embeddings and an Arcface model \cite{deng2019arcface} to extract identity embeddings. These two derived embeddings were then combined by a linear layer and then fed into the diffusion model to control the output identity.
InstantID \cite{wang2024instantid} introduced an IdentityNet to transfer the identity information from the reference to the input image. The IdentityNet added both semantic and spatial conditions induced from facial landmarks, images, and text prompts to guide the image generation process.

Unlike methods that primarily control internal facial appearances (e.g., color and shape of eyes, shape of nose, skin tone), our method focuses on controlling facial structures (e.g., contour, pose, shadows, highlights, and proportion of facial characteristics). Since internal facial appearances are also significant in depicting facial identity, our method can be combined with these approaches such as IP-Adapter \cite{ye2023ipadapter} for better ID preservation.

\section{Proposed Method}

The proposed NNSG-Diffusion is designed to adopt the closest reference face as guidance, allowing the downstream FAE process to introduce fewer changes to the input’s identity but ensuring improvements in facial aesthetics.
In this section, we first introduce preliminaries of the proposed method in Sec. \ref{sec:prelim}. Then we explain the designs of our three modules: NNFS module in Sec. \ref{sec:nnfs}, FGE module in Sec. \ref{sec:FGE}, and FB Sec. \ref{sec:fb}. Finally, we present the overall structure and workflow of our NNSG-Diffusion in Sec. \ref{sec:overview}.

\subsection{Preliminaries}
\label{sec:prelim}

3D Morphable Model (3DMM) \cite{blanz2023morphable,paysan20093d,deng2019accurate} is a fundamental model for facial reconstruction from 2D to 3D. The core concept of 3DMM is that faces can be represented as a linear combination of basis vectors in the 3D space. These basis vectors are orthogonal and parameterized, and allowed to flexibly and explicitly control the 3D facial attributes. 
Specifically, the reconstructed 3D face is defined by the facial shape $\boldsymbol{\mathrm{S}}\in\mathbb{R}^{3 N_v}$ and texture $\boldsymbol{\mathrm{T}}\in\mathbb{R}^{3 N_v}$ with $N_v$ vertices \cite{blanz2023morphable,deng2019accurate}:

\begin{figure}[t]
    \centering
    \includegraphics[width=1\linewidth]{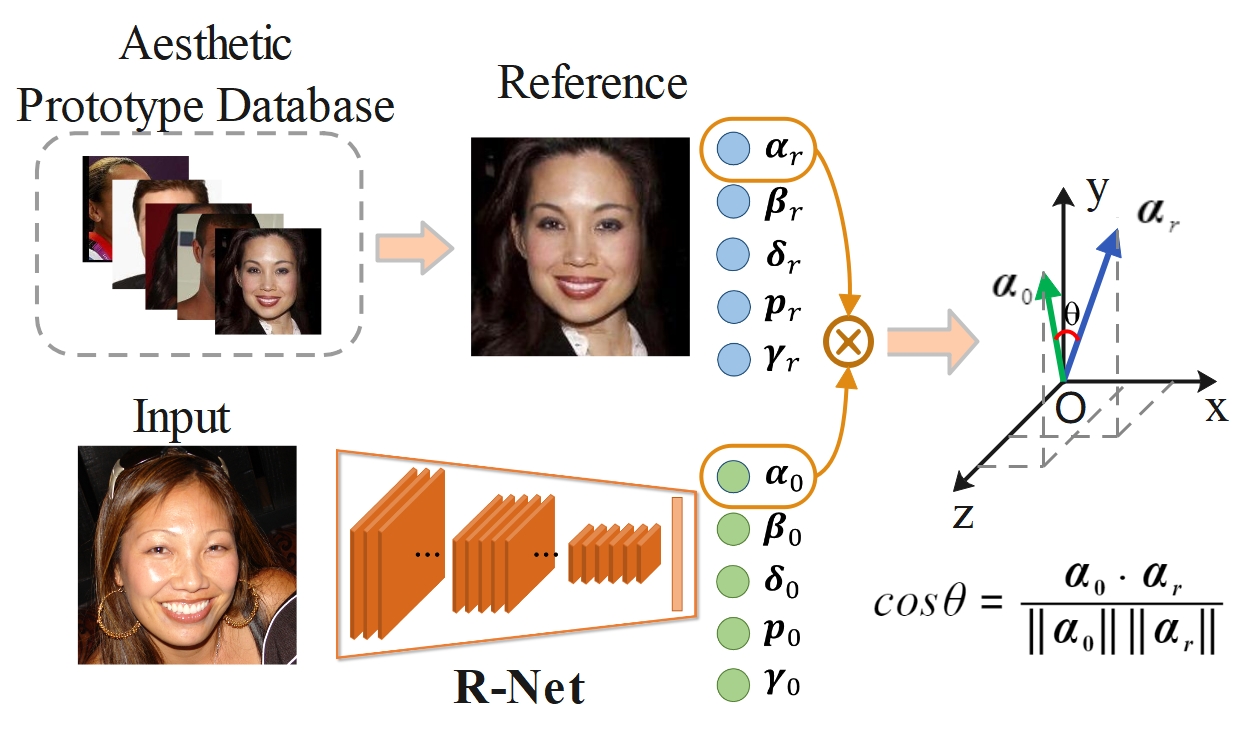}
    \caption{Design of the proposed NNFS module. R-Net \cite{deng2019accurate} is used to predict parameter vectors of decoupled 3D facial attributes: identity $\bm \alpha$, expression $\bm \beta$, texture $\bm{\delta}$, pose $\bm p$, and lighting $\bm \gamma$.}
    \label{Fig.3}
\end{figure}

\begin{equation}
\boldsymbol{\mathrm{S}}=\boldsymbol{\mathrm{S}}(\bm{\alpha},\bm{\beta}) =\bar{\boldsymbol{\mathrm{S}}}+\boldsymbol{\mathrm{B}}_{id}\bm{\alpha}+\boldsymbol{\mathrm{B}}_{exp}\bm{\beta}, 
\label{Eq.5}
\end{equation}
\begin{equation}
\boldsymbol{\mathrm{T}}=\boldsymbol{\mathrm{T}}(\bm{\delta})=\bar{\boldsymbol{\mathrm{T}}}+\boldsymbol{\mathrm{B}}_t\bm{\delta}, \label{Eq.6}
\end{equation}
where $\bar{\boldsymbol{\mathrm{S}}}$ and $\bar{\boldsymbol{\mathrm{T}}}$ denote the neutral facial shape and texture, $\boldsymbol{\mathrm{B}}_{id}$, $\boldsymbol{\mathrm{B}}_{exp}$, and $\boldsymbol{\mathrm{B}}_t$ are PCA bases of identity, expression, and texture; $\bm{\alpha}$, $\bm{\beta}$, and $\bm{\delta}$ are the corresponding parameter vectors for weights applying to the identity, expression and texture basis vectors for 3D face reconstruction. In our implementation, we adopt the deep learning-based and 3DMM-based method R-Net \cite{deng2019accurate} for the 3D face reconstruction, and details will be given in later sections.

\begin{figure}[t]
    \centering
    \includegraphics[width=1\linewidth]{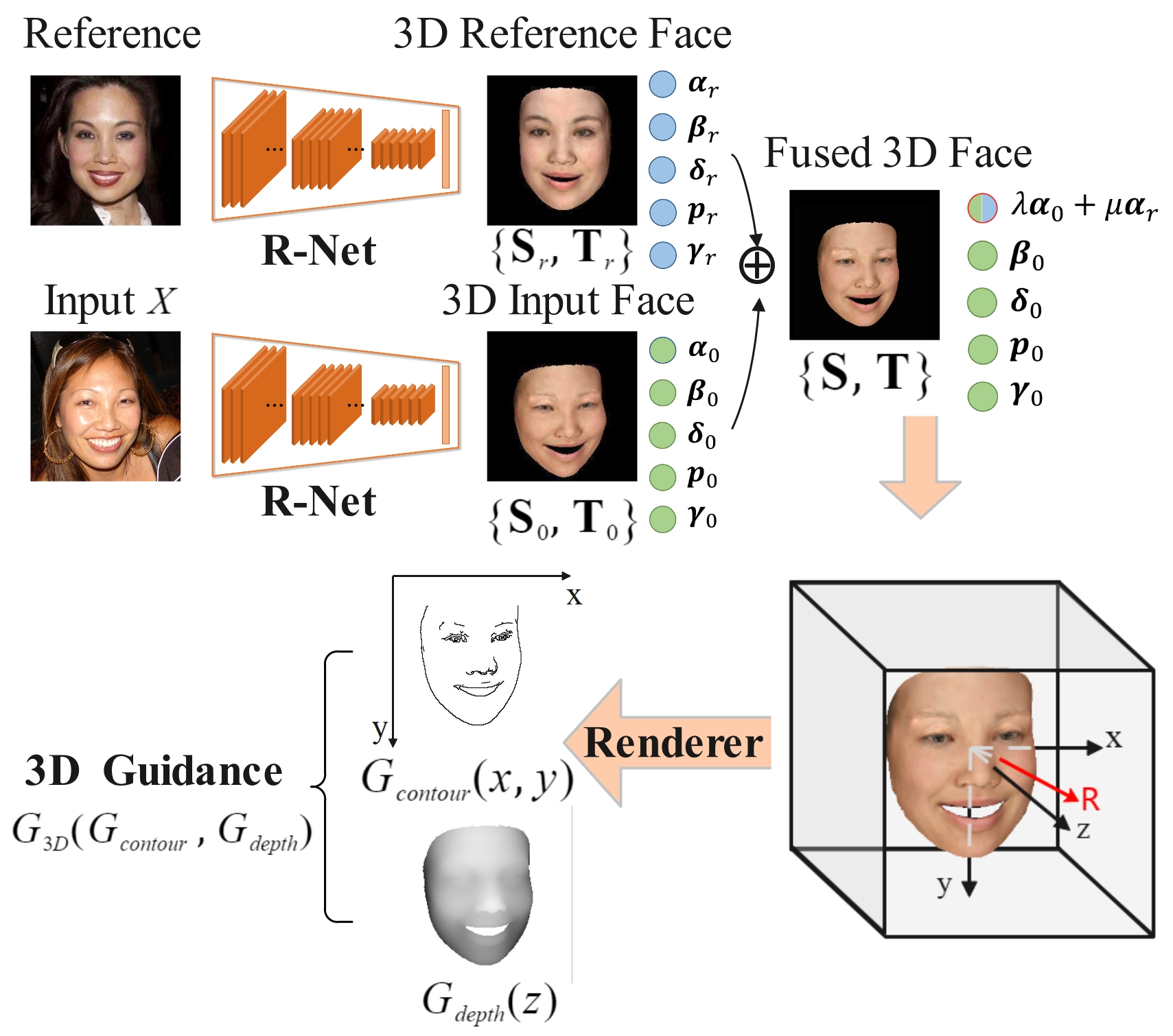}
    \caption{Design of the FGE module. FGE module projects the fused 3D face by the Pytorch3D Renderer for guidance extraction \cite{ravi2020pytorch3d, lassner2020pulsar}.}
    \label{fig:FGE_module}
\end{figure}

\begin{figure}[t]
    \centering
    \includegraphics[width=1\linewidth]{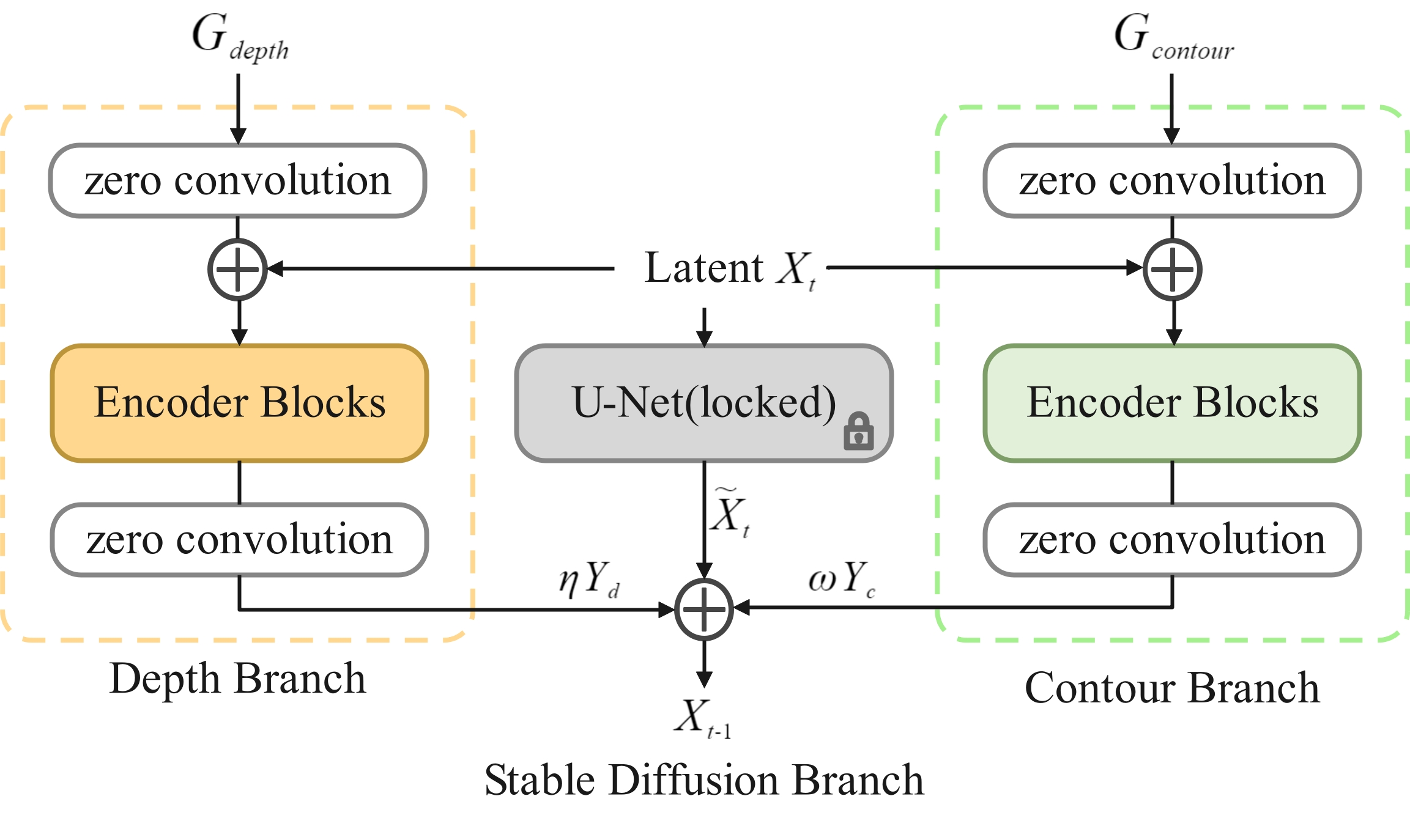}
    \caption{Demonstration of using depth and contour guidance in Stable Diffusion \cite{zhang2023adding} with control weights $(\omega, \eta)$. $Y_{c}$ and $Y_{d}$ are outputs of the intermediate block and encoding block and are directly fed into Stable Diffusion.}
    \label{fig:controlnet_module}
\end{figure}

\subsection{Proposed Method}

\subsubsection{Nearest Neighbor Face Searching (Module 1)}
\label{sec:nnfs}
The NNFS module aims to mitigate the loss of identity information of the input face brought by FAE by providing an appropriate reference. Thus, we search for an aesthetic prototype most similar to the input face in terms of 3D structure. 
Then we adopt the nearest aesthetic prototype as a reference face in the later FGE and FB modules.

\textbf{Aesthetic Prototype Database:} We first build a database to provide potential reference faces, so that we can find an aesthetic prototype (i.e., reference face with high attractiveness) most similar to the input face in terms of 3D structure. 
We utilize the CelebA dataset \cite{liu2015faceattributes} to establish a comprehensive aesthetic prototype database. The CelebA dataset contains 202,599 images, which include 10,178 celebrity faces with 40 annotated attributes for each image (e.g., ``Arched\_Eyebrows'', ``Mustache'', ``Black\_Hair'', and ``Eyeglasses''). In this study, we pick out the images with specific attributes (``attractive'' and ``young'') and remove images with low facial attractiveness or attributes (``bags under eyes'', ``Bald'', ``big nose'', ``blurry'', ``chubby'', ``double chin'', ``eyeglasses'', ``narrow eyes'', ``wearing hat'').
Subsequently, we select 6,610 faces with diverse identity attributes to provide reference faces. These facial images are composed of 4,797 female and 1,813 male facial images. Our database represents each image with parameter vectors of identity, expression, texture, pose, and lighting. These five attributes are decoupled by R-Net \cite{deng2019accurate} and saved for the use in later steps. 

\textbf{Face Searching:} The NNFS module (Fig. \ref{Fig.3}) adopts parameter vectors of decoupled facial attributes to search for the most similar face as the input one.
The main goal of decoupling facial attributes is to construct a latent space consisting of linear subspaces, where each linear subspace controls one facial attribute. Thus, we can manipulate or extract facial individual attributes by modifying corresponding latent variables or features in subspaces  \cite{karras2019style}. In this work, we use R-Net \cite{deng2019accurate} to predict parameter vectors of facial attributes: identity $\bm{\alpha}\in\mathbb{R}^{80}$, expression $\bm{\beta}\in\mathbb{R}^{64}$, texture $\bm{\delta}\in\mathbb{R}^{80}$, pose $\bm{p}\in\mathbb{R}^{6}$, and lighting $\bm{\gamma}\in\mathbb{R}^{9}$. Note that the backbone network of R-Net is ResNet-50. In particular, the last fully connected layer of the ResNet-50 is replaced by 239 neurons \cite{deng2019accurate}. Then the input face image $X_0\in\mathbb{R}^{3\times224\times224}$ is fed into the R-Net, resulting in a combined parameter vector of input face $x_0=(\bm{\alpha}_0, \bm{\beta}_0, \bm{\delta}_0, \bm{p}_0, \bm{\gamma}_0)\in\mathbb{R}^{239}$. 
Denoting the identity parameters of a reference face in the Aesthetic Prototype Database with $\bm{\alpha}_r$ (pre-computed by R-Net when the database was built), we calculate the cosine similarity between $\bm{\alpha}_0$ and $\bm{\alpha}_r$ to search the nearest aesthetic prototype, as denoted by Eq. (\ref{Eq.3}) as:

 \begin{equation}
    cos\theta = \frac{\bm{\alpha}_0 \cdot \bm{\alpha}_r }{\Vert\bm{\alpha}_0\Vert  \Vert\bm{\alpha}_r\Vert}.
    \label{Eq.3}
\end{equation}
Note that here we mainly use the identity parameter vector $\bm{\alpha}$ in 3DMM for extracting the 3D structure guidance, to avoid expressions to be transferred in FAE guidance. 
We do not consider texture parameters $\bm{\delta}$ in the 3D structure guidance, since such texture feature can be better represented by 2D features in the case of FAE features.

A higher value of cosine similarity shows a closer structural distance between the aesthetic prototype and the input face. The NNFS module can be formulated as Eq. (\ref{Eq.4}):
 
\begin{equation}
    \mathcal{F}_{NNFS}(\bm{\alpha}_0)=\underset{j}{\mathrm{argmax}}(\{cos(\bm{\alpha}_0, \bm{\alpha}_j)| \bm{\alpha}_j \in \mathcal{V}^N\}), \label{Eq.4}
\end{equation}
where $\mathcal{V}^N=\{\bm{\alpha}_j \vert j \in [1, N], \bm{\alpha}_j\in\mathbb{R}^{80}\}$ is the vector space consisting of 3D identity feature vectors of our database, and $N$ is the number of images in the database. By calculating the cosine similarity, the NNFS module $\mathcal{F}_{NNFS}(\cdot)$ can search one identity parameter vector $\bm{\alpha}_j$ resulting in the maximum value of cosine similarity with $\bm{\alpha}_0$, where $j$ is the order number of the nearest aesthetic prototype in the Aesthetic Prototype Database.
Then the structure and texture representation $\{\boldsymbol{\mathrm{S}}_j\in\mathbb{R}^{3 N_v}, \boldsymbol{\mathrm{T}}_j\in\mathbb{R}^{3 N_v}\}$ can be derived according to the order number of the matched face from saved parameters of our Aesthetic Prototype Database.
The examples of search results are shown in Fig. \ref{fig:Module1_2_Comparative_experiment}(A). Our NNFS module selects the face with the nearest structure as the aesthetic prototype, providing a reference for subsequent facial enhancement. 
This enables our method to introduce fewer identity information loss caused by facial deformation. 
Additionally, since we use the identity parameter vector $\bm{\alpha}$ instead of directly utilizing the facial structure vector $\boldsymbol{\mathrm{S}}$ (in Eq. (\ref{Eq.5})) for face searching, the influence of facial expressions can be mitigated.

\begin{figure*}[t]
    \centering
    \includegraphics[width=1\linewidth]{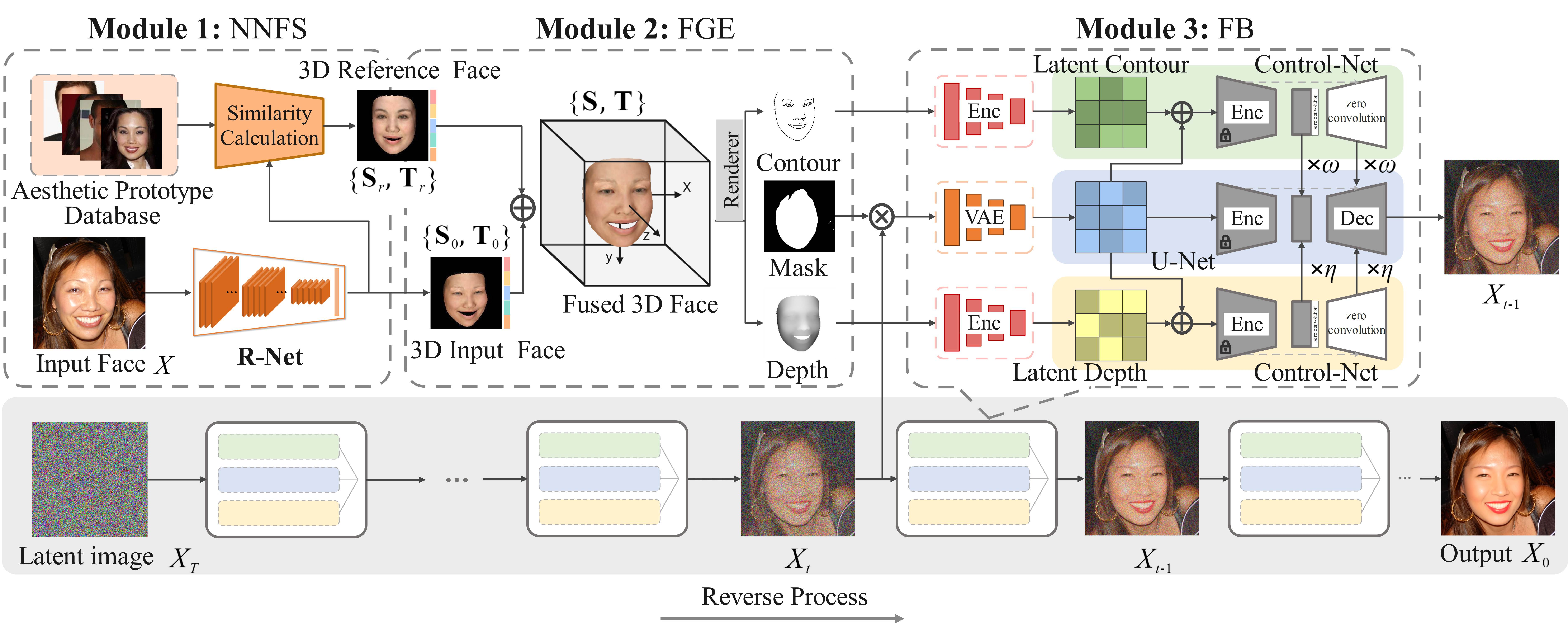}
    \caption{The framework of proposed Nearest Neighbor Structure Guidance Based on Diffusion Model (NNSG-Diffusion). It consists of three modules: Nearest Neighbor Face Searching (NNFS), Facial Guidance Extraction (FGE), and Face Beautification (FB). Our method adopts the NNFS module to search reference face most similar to the input face, and fuses reference and input faces to extract 3D structure guidance by FGE module. Finally, the FB module can accurately control the process of beautification by using the 3D structure guidance.}
    \label{fig:framwork_new}
\end{figure*}

\vspace{3mm}
\subsubsection{Facial Guidance Extraction (Module 2)}
\label{sec:FGE}

This module aims to produce FAE guidance with parameter vectors of reconstructed 3D face models of input and reference face. 
The FGE module first fuses the identity parameters of the input face and the reference face by a weighted sum. 
To extract guidance, the fused 3D face is projected in the view plane of the input face by the Pytorch3D Renderer \cite{ravi2020pytorch3d,lassner2020pulsar}.

As shown in Fig. \ref{fig:FGE_module}, we only edit the identity parameter of the 3D model of input image and do not change other attributes. 
This allows the 3D guidance produced from the reference image to provide guidance only about the face structure, regardless of what the facial expressions, poses or lighting conditions of the reference face are.
After the fusion procedure, the fused identity parameter can be represented by:
\begin{equation}
    \bm{\alpha} = \lambda  \bm{\alpha}_0 + \mu  \bm{\alpha}_r,
\end{equation}
where $\lambda$ and $\mu$ are the weight parameters, and both of them are set to 0.5. $\bm{\alpha}_r$ is the identity parameter vector of the reference face searched by NNFS module.

According to Eq. (\ref{Eq.5}), we assume that the identity parameter is the most influential parameter to facial identities. Previous FAE methods \mbox{\cite{chen2023aep, liu2022face, he2020fagans}} usually indirectly consider the preservation of identity during the training stage via identity loss, while such constraints may not be effective in the actual inference stage for FAE.
To preserve facial identity consistency, our method directly adopts the weighted sum method of the identity parameters of the input face and the reference face during the inference stage of FAE, which helps preserve a portion of the original identity features. We fuse the facial attributes of input and reference by a linear combination of feature vectors, as shown in Fig. \ref{fig:FGE_module}. As \cite{deng2019accurate}, we use the perspective camera model with an empirically selected focal length for the 3D to 2D projection structure. The fused 3D face pose $p$ consists of rotation $\boldsymbol{\mathrm{R}}\in \mathrm{SO}(3)$ and translation $\boldsymbol{\mathrm{t}}\in \mathbb{R}^3$.  Thus, we have the rendered 3D face model for the 3D guidance extraction as: 
\begin{equation}
    M_{face} = \mathcal{F}_{renderer}(\{\boldsymbol{\mathrm{S}}_0, \boldsymbol{\mathrm{T}}_0\}, \{\boldsymbol{\mathrm{S}}_r, \boldsymbol{\mathrm{T}}_r\}, \boldsymbol{\mathrm{R}}, \boldsymbol{\mathrm{t}}),
    \label{Eq.10}
\end{equation}
where a 3D face model $M_{face}$ is rendered from the given parameter sets: $\{\boldsymbol{\mathrm{S}}_0, \boldsymbol{\mathrm{T}}_0\}, \{\boldsymbol{\mathrm{S}}_r, \boldsymbol{\mathrm{T}}_r\}, \boldsymbol{\mathrm{R}}, \boldsymbol{\mathrm{t}}$. $\{\boldsymbol{\mathrm{S}}_0\in\mathbb{R}^{3 N_v}, \boldsymbol{\mathrm{T}}_0\in\mathbb{R}^{3 N_v}\}$ denotes the 3D representation of the input face, and $\{\boldsymbol{\mathrm{S}}_r\in\mathbb{R}^{3 N_v}, \boldsymbol{\mathrm{T}}_r\in\mathbb{R}^{3 N_v}\}$ is the 3D representation of the reference face.

\begin{figure}[t]
    \centering
    \includegraphics[width=1\linewidth]{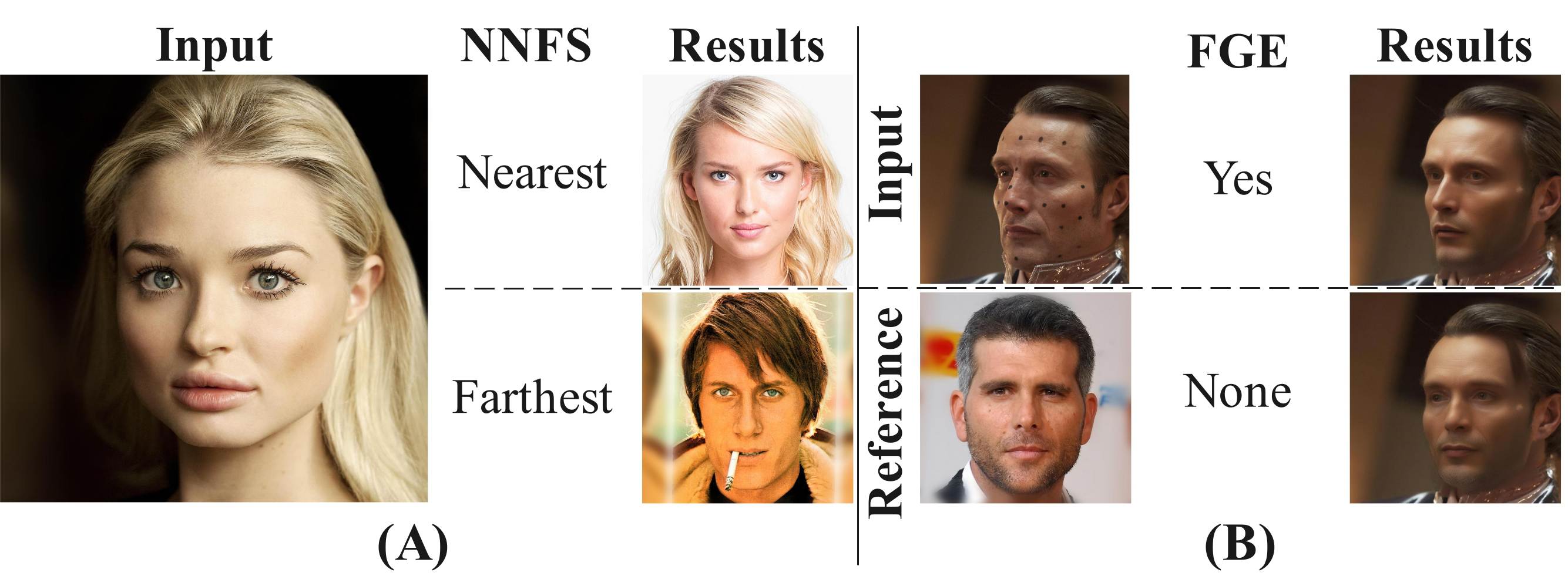}
    \caption{(A) Results of the nearest and the farthest aesthetic prototypes by matched by NNFS module; (B) The comparison between the results with the FGE module and without FGE module, where the pose of the resulting face can be different from the input one when the proposed FGE module is not employed.}
    \label{fig:Module1_2_Comparative_experiment}
\end{figure}

\begin{figure}[t]
    \centering
    \includegraphics[width=1\linewidth]{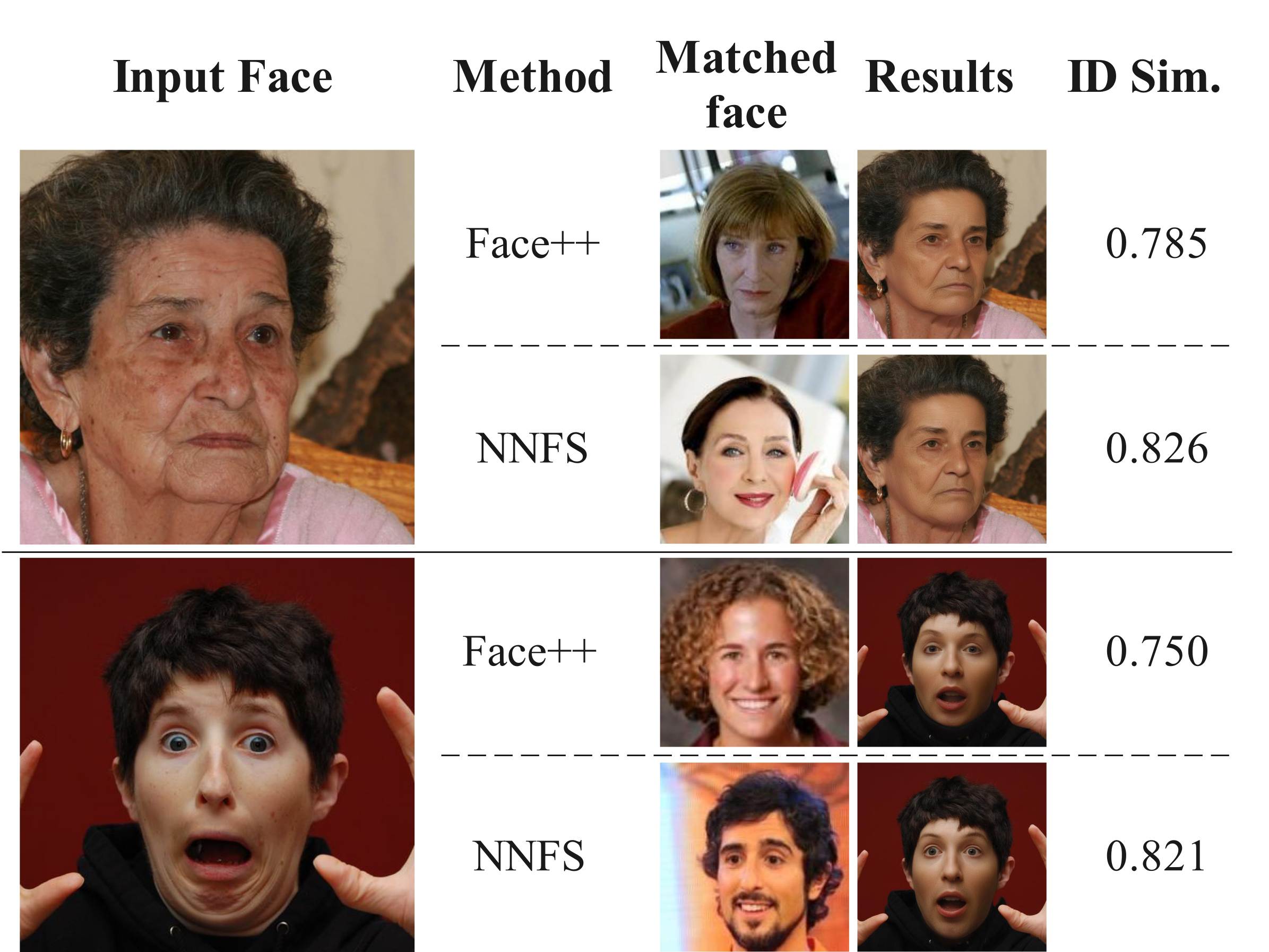}
    \caption{Comparing different face searching methods in FAE. These examples show that the NNFS module searches for faces with the nearest 3D facial structures and retains more 3D structural identity.}
    \label{Fig.7}
\end{figure}

In this work, Pytorch3D Renderer \cite{ravi2020pytorch3d,lassner2020pulsar} is used to project the 3D face to obtain the 3D structure guidance $G_{3D}$ from the fused 3D face model. The guidance $G_{3D}$ involves two parts: the contour guidance and the depth guidance. 
The contour guidance $\mathrm{G}_{contour}$ represents the contours of the face on the view plane and is obtained by applying a canny edge operator to the 2D projection of the 3D face: $G_{contour} = \mathcal{F}_{canny} \circ \mathcal{F}_{project}(M_{face})$. 
The depth guidance $\mathrm{G}_{depth}$ represents the face's depth information in the direction of the ray, and it is extracted from the rendered 3D face, denoted as $G_{depth} = \mathcal{F}_{depth}(M_{face})$. Thus, the 3D structure guidance $G_{3D} = \{G_{depth}, G_{contour}\}$ is composed of the contour and the depth guidance, and $\mathcal{F}_{FGE}(\cdot)$ represents the FGE module, and $M_{face}$ represents reconstructed 3D face model.
In summary, 3D structure guidance can be expressed as:
\begin{equation}
    G_{3D}=\mathcal{F}_{FGE}(M_{face}).
    \label{Eq.11}
\end{equation}

\subsubsection{Face Beautification (Module 3)}
\label{sec:fb}
The beautification process is implemented by SD \cite{rombach2022high} with ControlNet \cite{zhang2023adding}. 
The overall design of the FB module is demonstrated in Fig. \ref{fig:controlnet_module}.
In the FB module, we mask the face region of the input image, and adopt local inpainting of SD to fill the masked region. 
In particular, the area of the face mask is slightly larger than the input face, so that the part generated by SD can transit to the original background more naturally \cite{couairon2022diffedit}. 

The depth and the contour guidance are used as extra inputs to ControlNet \cite{zhang2023adding}, and the weights of the U-Net of SD branch are frozen. 
The U-Net of SD is defined as $\mathcal{C}(\cdot)$ parameterized by $\Theta$, which remains fixed during encoding. If $X_t$ is considered as the latent spatial feature at the $t$-th step in the reverse process, the corresponding denoised feature  $\tilde{X_t}$ can be obtained by $\tilde{X_t}=\mathcal{C}(X_t;\Theta)$. The U-Net is connected to the ControlNet branch through the ``zero convolution'' layer, and the zero convolution operation is represented by $\mathcal{H}(\cdot;\cdot)$. $\left\{\Theta_{1}, \Theta_{2}\right\}$ represent the parameters of two ``zero convolution'' layers and $\Theta'$ represents the parameter of the trainable copy in ControlNet. Hence, the output feature $Y$ can be represented as follows \cite{zhang2023adding}:
\begin{equation}
\label{eq:controlnet}
    Y = \mathcal{C}(X_t;\Theta) + \mathcal{H}(\mathcal{C}(X_t + \mathcal{H}(G_*;\Theta_{1});\Theta');\Theta_{2}),
\end{equation}
where $G_*$ is either a canny edge map (contour) or a depth map as the guidance.

The output features of the contour and depth branches are denoted as $\left\{Y_{c}, Y_{d}\right\}$. The output of the $t-1$ step in the reverse process of stable diffusion enhanced by contour and depth clues can be represented as follows:
\begin{equation}
    X_{t-1} = \tilde{X_t} + \omega Y_c + \eta Y_d,
\label{eq:guidance_fusion}
\end{equation}
where the weight parameters $(\omega, \eta)$ are used to balance the effect of each branch on the image generation process. 

\subsection{Overall Framework}
\label{sec:overview}
In summary, the overall framework of the proposed NNSG-Diffusion is shown in Fig. \ref{fig:framwork_new}. The output image $X_0$ of our FAE method is given as: 
\begin{equation}
\label{eq:ffb1}
X_0=\mathcal{F}_{FAE}(X, {G}_{3D}),
\end{equation}
where $X$ is the input image and ${G}_{3D}$ expresses the 3D structure guidance.
The 3D structure guidance $G_{3D}$ in Eq. (\ref{eq:ffb1}) is extracted by the FGE module
$\mathcal{F}_{FGE}(\cdot)$: 
\begin{equation}
X_0=\mathcal{F}_{FAE}(X, \mathcal{F}_{FGE}(\mathcal{F}_{renderer}(\{\boldsymbol{\mathrm{S}}_0, \boldsymbol{\mathrm{T}}_0\}, \{\boldsymbol{\mathrm{S}}_r, \boldsymbol{\mathrm{T}}_r\}, \boldsymbol{\mathrm{R}}, \boldsymbol{\mathrm{t}}))), 
\end{equation}
where $\{\boldsymbol{\mathrm{S}}_0\in\mathbb{R}^{3 N_v}, \boldsymbol{\mathrm{T}}_0\in\mathbb{R}^{3 N_v}\}$ denotes the 3D representation of the input face, $\{\boldsymbol{\mathrm{S}}_r\in\mathbb{R}^{3 N_v}, \boldsymbol{\mathrm{T}}_r\in\mathbb{R}^{3 N_v}\}$ is the 3D representation of the reference face, which is matched by NNFS module $\mathcal{F}_{NNFS}(\cdot)$.  $\boldsymbol{\mathrm{R}}\in \mathrm{SO}(3)$ denotes the input face's rotation and $\boldsymbol{\mathrm{t}}\in \mathbb{R}^3$ denotes the translation. 
$N_v$ is the number of vertices of the 3D face model. 
NNFS module $\mathcal{F}_{NNFS}(\cdot)$ searches the nearest structure aesthetic prototype by calculating cosine similarity (Eq. (\ref{Eq.3})) and obtains parameter vectors of the decoupled facial attributes of the aesthetic prototype. These facial attribute vectors are applied to reconstruct the structure and texture features $\{\boldsymbol{\mathrm{S}}_r, \boldsymbol{\mathrm{T}}_r\}$ of reference 3D face. To extract the facial depth $G_{depth}$ and contour $G_{contour}$ guidances, we use the PyTorch3D Renderer \cite{ravi2020pytorch3d,lassner2020pulsar} for the 3D to 2D projection structure by a Camera Model \cite{deng2019accurate} in FGE module $\mathcal{F}_{FGE}(\cdot)$. Facial depth $G_{depth}$ and contour $G_{contour}$ guidance jointly compose the 3D structure guidance ${G}_{3D}$. Finally, the FB module extracts both depth and contour features with ControlNet as Eq. (\ref{eq:controlnet}) to enhance the input feature $X$ to produce the final results.

\section{EXPERIMENTS}
\subsection{Experimental Setup}

\begin{figure}[t]
    \centering
    \includegraphics[width=1\linewidth]{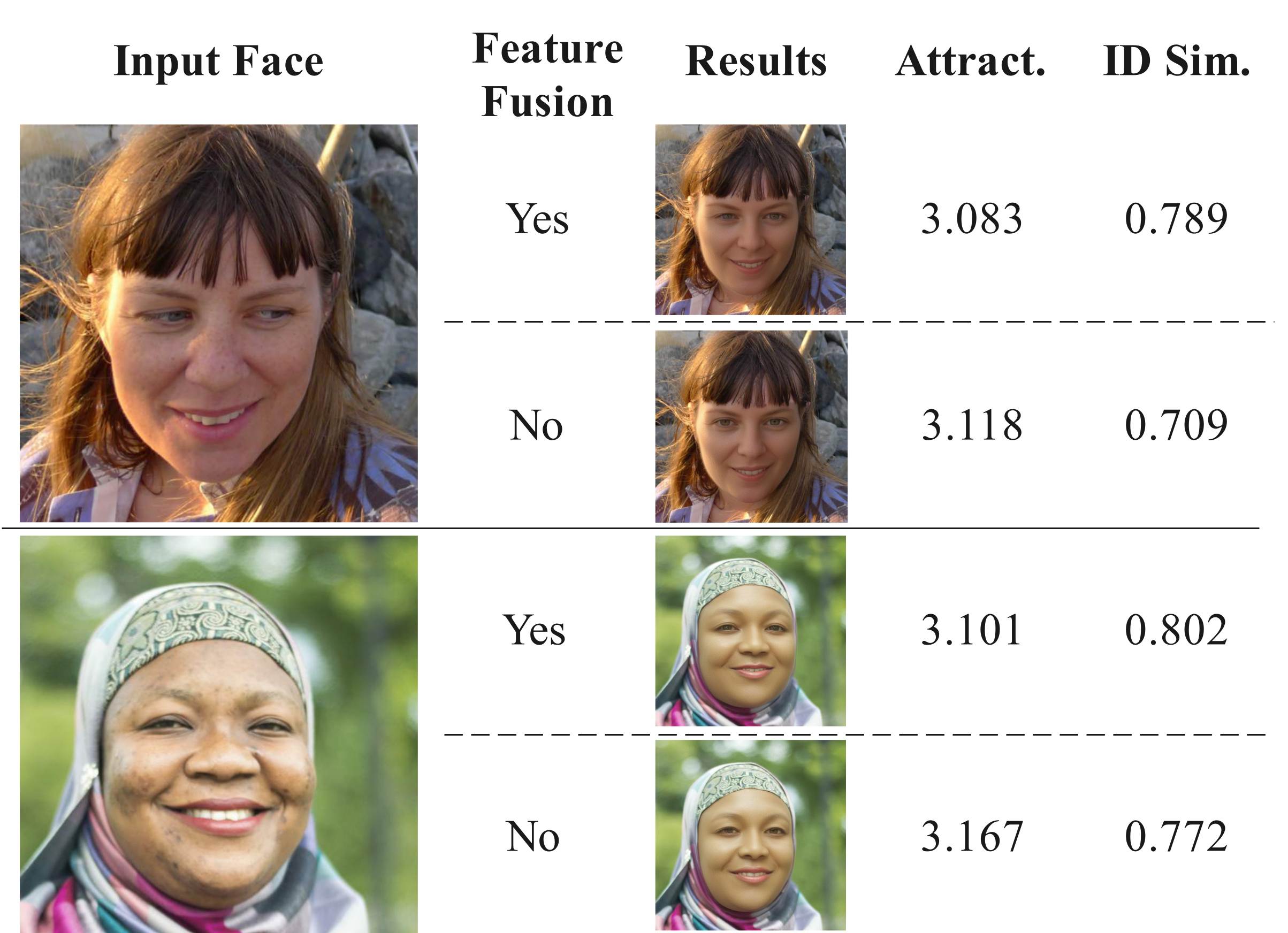}
    \caption{Comparing results with facial feature fusion and that without feature fusion. Results demonstrate that facial structure deformation is more pronounced in the absence of feature fusion, leading to a decrease of ID Similarity.}
    \label{Fig.8}
\end{figure}

\begin{table}[t]
\caption{Performance of beauty evaluation models.}
\label{tab:beauty_evaluation_model}
\begin{center}
\begin{tabular}{lccc}
\toprule[1pt]
Method   & PC     & RMSE  & MAE   \\ \midrule[0.5pt]
VGG16-based    & 0.812 & 0.555 & 0.414 \\
ResNet18-based & \textbf{0.915}  & \textbf{0.275} & \textbf{0.210} \\ \bottomrule[1pt]
\end{tabular}
\end{center}
\end{table}

\textbf{Evaluation Metrics:} we mainly adopt the following metrics for our evaluation:

\textit{1) FID:} The Fréchet Inception Distance (FID) \cite{heusel2017gans} serves as an indicator of the quality of generated images across different methods. It is calculated by computing the Fréchet distance between two Gaussian distributions of input and generation images fitted by features from the Inception network \cite{szegedy2017inception}. A lower FID value corresponds to higher sample quality.

\textit{2) PSNR:} Peak Signal-to-Noise Ratio (PSNR) \cite{yuanji2003image} is an image quality metric that assesses the error between corresponding pixel points of the generated image and input image. Based on Mean Squared Error (MSE), a higher PSNR value indicates superior image quality.

\textit{3) SSIM:} Structural Similarity (SSIM) \cite{wang2004image} gauges the structural resemblance between the generated image and input image, aligning more closely with human visual perception compared to PSNR. SSIM values range from 0 to 1, with larger values signifying higher similarity.

\textit{4) Attractiveness:} We employ beauty evaluation models based on VGG16 \cite{diamant2019beholder} and ResNet18 \cite{2018SCUT} trained on SCUT-5500 dataset \cite{2018SCUT} for evaluating the attractiveness of generated images. 60\% of the SCUT-5500 dataset is used for training and the rest 40\% is used for testing.
As most existing perceptual quality assessment methods that predict subjective viewing scores \cite{2018SCUT,hou2023towards,wu2023neighbourhood}, to evaluate the effectiveness of the adopted attractiveness evaluation models, we adopt Pearson correlation (PC), root mean squared error (RMSE) and mean squared error (MSE) to evaluate the consistency between the predicted scores and GT subjective scores (i.e., attractiveness in our case).
For the VGG16-based model, the test results are 0.812, 0.555, and 0.414 for PC, RMSE, and MAE, respectively. For the ResNet18-based model, the test results are 0.915, 0.275, and 0.210 for PC, RMSE, and MAE, respectively.
We have included these test results in Table \ref{tab:beauty_evaluation_model} to demonstrate the effectiveness of our evaluation model. 
The adopted attractiveness evaluator produces scores range from 1 to 5, with larger values signifying greater attractiveness. 
The Attractiveness is denoted as ``Attract.'' for short in later tables and figures.

\begin{table*}
\caption{Quantitative analysis results on FFHQ dataset comparing the use of different modules of the proposed method. The upward arrows indicate that larger values of the evaluation metric correspond to better results, while the downward arrows indicate the opposite. The used version of Stable Diffusion model is SDXL-1.0 (with IP-Adapter). The mean Attractiveness scores for the inputs are 2.325 with VGG16 and 2.006 with ResNet18.}
\begin{center}
\begin{tabular}{ccccccccccc}
\toprule[1pt]
\multicolumn{2}{c}{Searching} & \multirow{2}{*}{Feature Fusion} & \multicolumn{2}{c}{Guidance}   & \multicolumn{2}{c}{Attractiveness$\uparrow$} & \multirow{2}{*}{ID Similiarity$\uparrow$} & \multirow{2}{*}{FID$\downarrow$} & \multirow{2}{*}{PSNR$\uparrow$} & \multirow{2}{*}{SSIM$\uparrow$} \\ \cmidrule(r){1-2} \cmidrule(lr){4-5} \cmidrule(lr){6-7}
Face++         & NNFS         &                                 & Contour        & Depth         & VGG16          & ResNet18          &                                 &                      &                       &                       \\ \midrule[0.5pt]
$\checkmark$   &              & $\checkmark$                    & $\checkmark$   & $\checkmark$  & 2.994          & 2.658             & 0.750                           & 34.544               & 22.616                & 0.795                 \\
               & $\checkmark$ &                                 & $\checkmark$   &  $\checkmark$ & 3.063          & 2.620             & 0.748                           & 39.049               & 21.857                & 0.783                 \\
               & $\checkmark$ & $\checkmark$                    &                &               & 3.013          & 2.724             & 0.766                           & 27.671               & 21.335                & 0.787                 \\
               & $\checkmark$ & $\checkmark$                    & $\checkmark$   &               & \textbf{3.108} & \textbf{2.781}    & 0.765                           & 30.908               & 21.651                & 0.782                 \\
               & $\checkmark$ & $\checkmark$                    &                &  $\checkmark$ & 2.965          & 2.633             & 0.790                           & \textbf{25.221}               & 22.637                & 0.800                 \\
               & $\checkmark$ & $\checkmark$                    & $\checkmark$   &  $\checkmark$ & 3.034          & 2.664             & \textbf{0.794}                  & 26.535      & \textbf{22.714}       & \textbf{0.801}        \\ \bottomrule[1pt]
\end{tabular}
\end{center}
\label{tab:table1_new}
\end{table*}

\textit{5) ID Similarity:} Maintaining identity consistency is a primary challenge for FAE methods. To address this, we calculate the cosine similarity between the facial identity feature ($\bm{\alpha}$ in Eq.(\ref{Eq.5})) of the beautified face and the input face. We use R-Net \mbox{\cite{deng2019accurate}}\footnote{\url{https://github.com/sicxu/Deep3DFaceRecon_pytorch}} to regress facial identity features. 
A higher ID Similarity value indicates less loss of facial identity information in terms of 3D structure. The value of ID Similarity ranges from 0 to 1, with larger values signifying a more similar facial structure. 
The ID Similarity is denoted as ``ID Sim.''  for short in later tables and figures.

\textbf{Evaluation Datasets:} Our evaluation is conducted on subsets collected from publicly available datasets. Specifically, we constructed four subsets by varying the data sources and sampling methods. 
The four evaluation datasets are FFHQ, CelebAMask-Hq (Mini), CelebAMask-Hq (Large), and SCUT-FBP. The details of data collection are given below:

\begin{itemize}
  \item CelebAMask-Hq (Mini): We select images with attractiveness scores lower than 3 (deemed as low attractiveness) evaluated by the employed beauty evaluation model \cite{diamant2019beholder}. This results in 169 images sampled from the CelebAMask-Hq dataset \cite{CelebAMask-HQ}.
  \item CelebAMask-Hq (Large): We exclude face images in CelebAMask-Hq dataset \cite{CelebAMask-HQ} occluded or labeled as `attractive' to construct the CelebAMask-Hq (Large) test set. This results in 6200 facial images for testing.
  \item FFHQ: We select 121 test images from the FFHQ dataset \cite{kazemi2014one} according to the low aesthetic scores (attractiveness score is less than 3) which are predicted by the employed beauty evaluation model \cite{diamant2019beholder}.
  \item SCUT-FBP: The SCUT-FBP dataset \cite{2018SCUT} contains 5500 facial images, and we use all of its facial images as our test data.
\end{itemize}

\begin{figure*}
    \centering
    \includegraphics[width=0.9\linewidth]{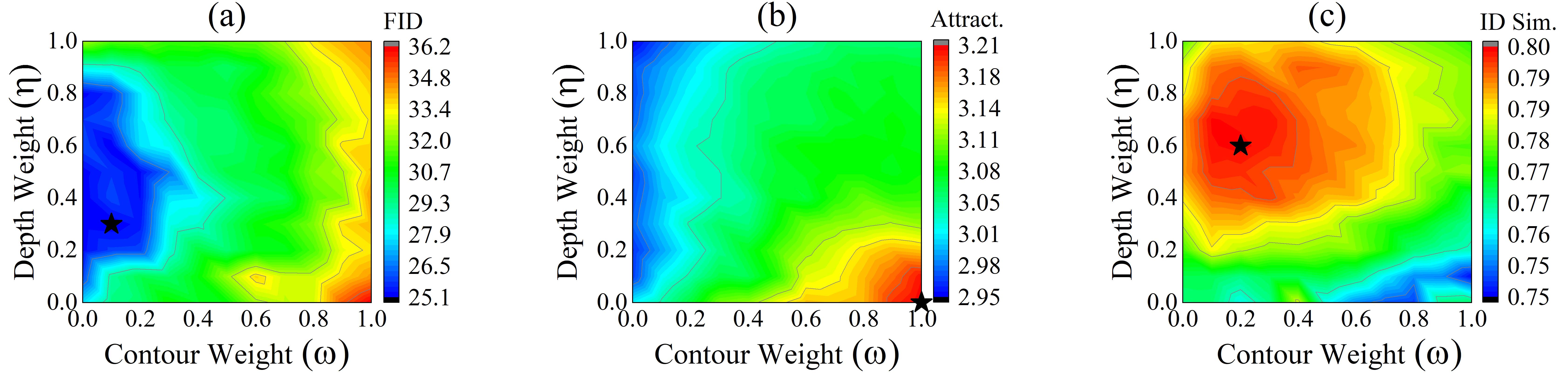}
    \caption{Demonstration of effects of weights on contour and depth guidance with NNSG\_SDXL + IP-Adapter on FFHQ dataset. 
    }
    \label{fig:para_for_depth_contour}
\end{figure*}

\begin{figure}
    \centering
    \includegraphics[width=1\linewidth]{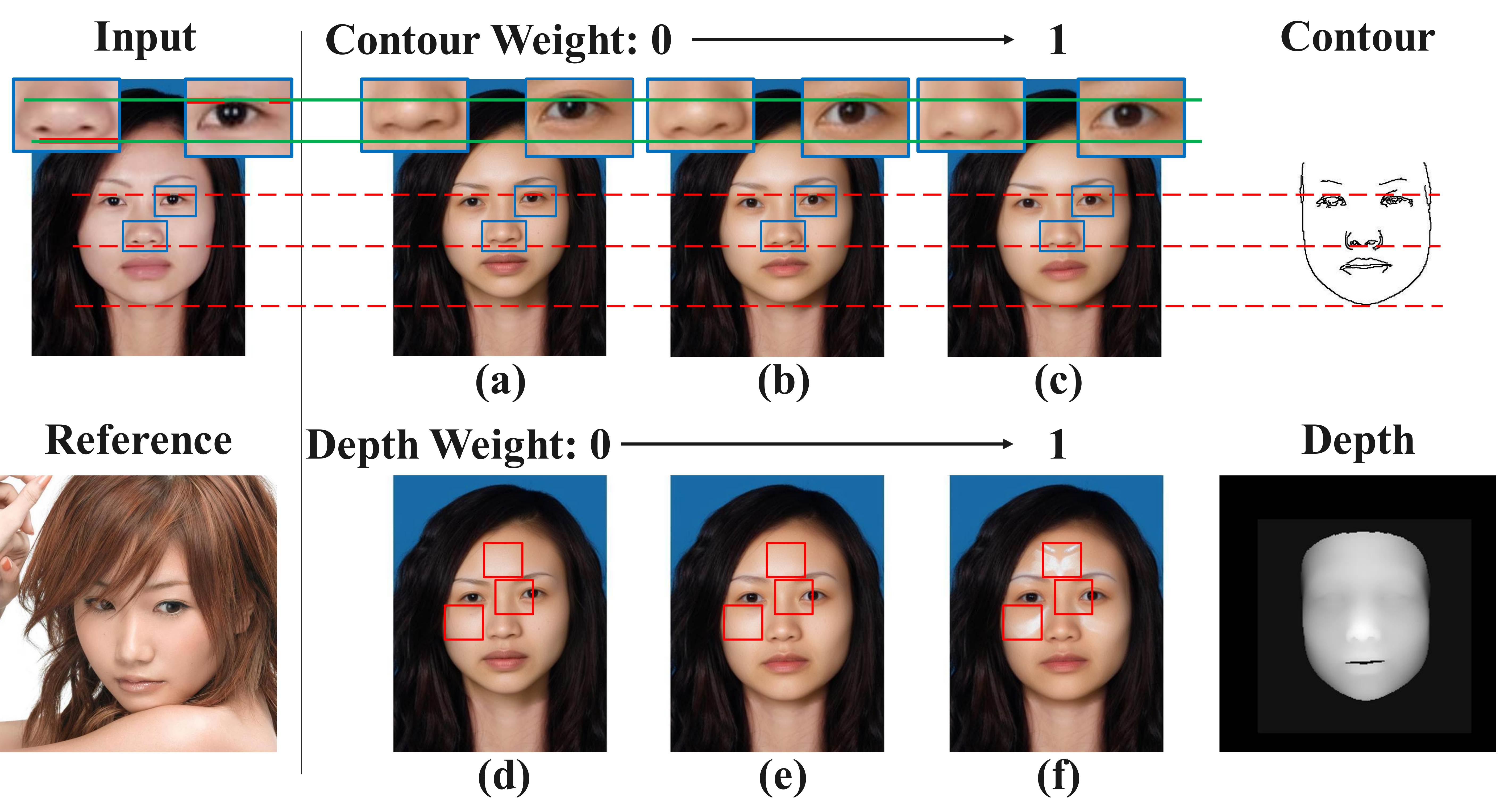}
    \caption{Demonstration of effects of contour and depth guidance in facial aesthetics enhancement process.}
    \label{fig:para_adject}
\end{figure}

\begin{figure}[t]
    \centering
    \includegraphics[width=1\linewidth]{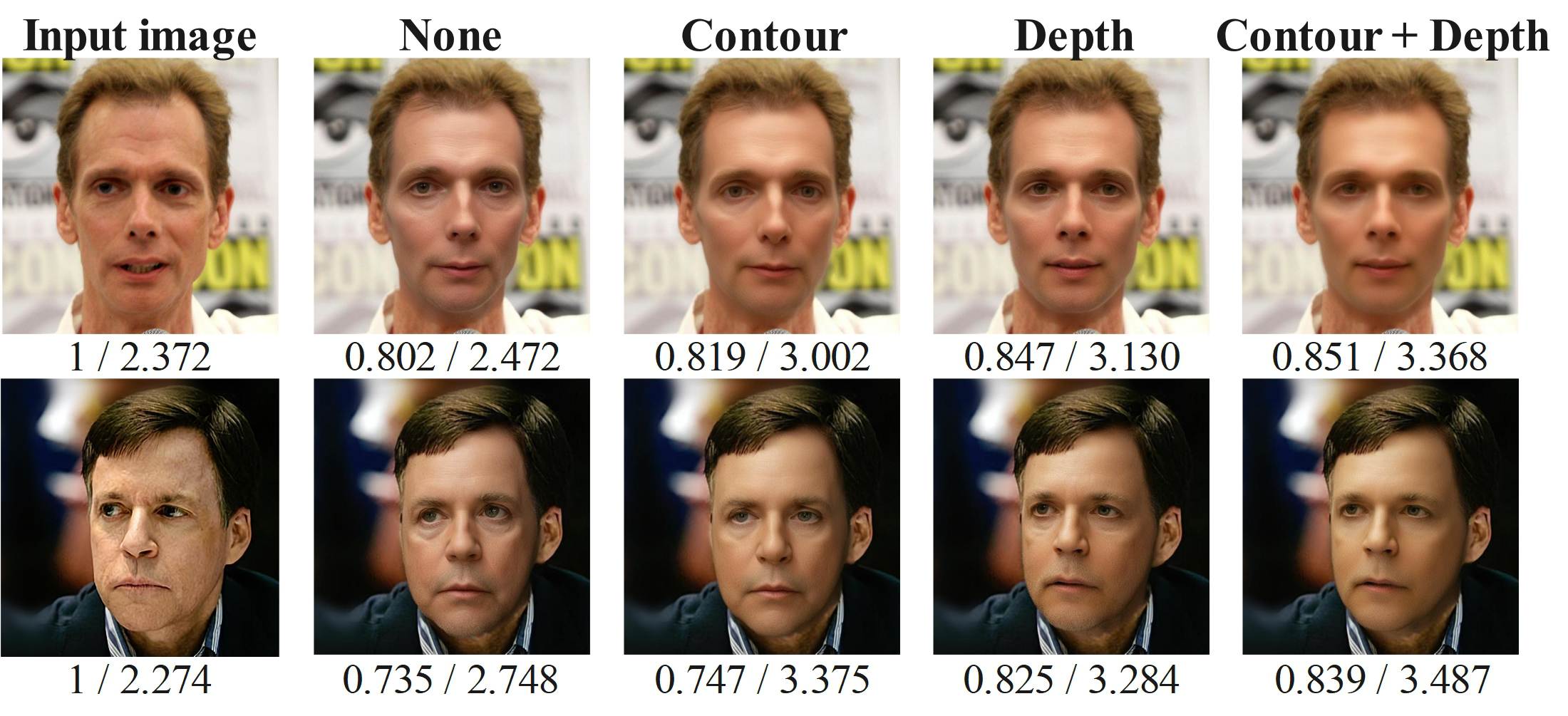}
    \caption{Demonstration of adding contour or depth guidance results in a more realistic face. The scores below each image are ID Similarity / Attractiveness.}
    \label{fig:exp_3d_structure}
\end{figure}

\textbf{Implementation Details:} 
The input 2D image is preprocessed by automatic alignment and cropping, resulting in the input size of $224 \times 224$. Significantly, we conducted comparative experiments with two versions of the SD model. The SD model versions used are SD-V1-2\footnote{\url{https://huggingface.co/runwayml/stable-diffusion-inpainting}} and SDXL-1.0\footnote{\url{https://stablediffusionxl.com}}, respectively.
The text prompt for the SD-V1-2 version is: ``AW photo, Attractive face, (high detailed skin: 1.2), soft lighting, high quality'', and the text prompt for SDXL-1.0 version is: ``breathtaking $\langle$ prompt $\rangle$, award-winning, portrait of human image, professional, highly detailed''. 
For both methods, we keep most parameters as default.
SDXL-1.0 is used with adjusted output size 1024×1024, number of inference steps 30, and guidance scale 1. SD-V1-2 is used with adjusted output size 1024×1024, number of inference steps 25, and guidance scale 4.
Meanwhile, we use the open-sourced model ControlNet (depth\footnote{\url{https://huggingface.co/lllyasviel/sd-controlnet-depth}} \footnote{\url{https://huggingface.co/diffusers/controlnet-depth-sdxl-1.0}}  and canny\footnote{ \url{https://huggingface.co/lllyasviel/sd-controlnet-canny}} \footnote{\url{https://huggingface.co/diffusers/controlnet-canny-sdxl-1.0}}) without training it. The weight parameter on contour and depth guidance in Eq. (\ref{eq:guidance_fusion}) is set to 0.3 and 0.5.

\begin{figure*}
    \centering
    \includegraphics[width=1\linewidth]{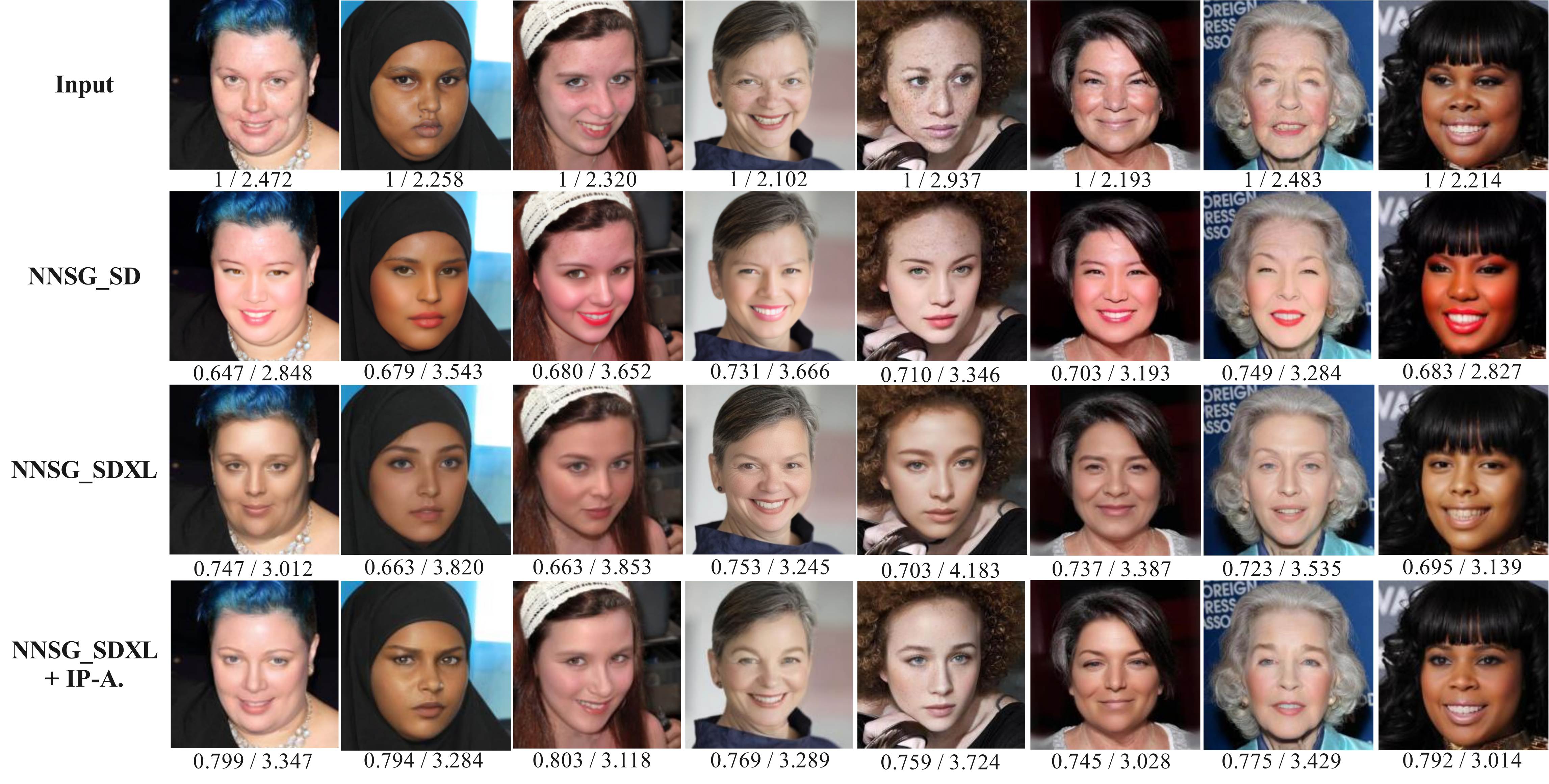}
    \caption{Comapring different versions of Stable Diffusion (SD) model. SDXL-1.0 version presents fewer issues about unnatural facial appearance generation (such as the overrepresented lip color, skin tone, and pupil color). IP-A. stands for IP-Adapter. The scores below each image are ID Similarity / Attractiveness.}
    \label{fig:NNSG_Version_comparison}
\end{figure*}

\subsection{Effectiveness of 3D Facial Feature Searching}
The proposed NNFS module is designed for face searching based on 3D face identity feature, as introduced in Sec. \ref{sec:nnfs}. 
To match a reference face, the identity feature of the input face is decoupled from the facial expression. Then, the cosine similarity between the input face and an aesthetic prototype is calculated based on the facial identity features.
In Fig. \ref{fig:Module1_2_Comparative_experiment}(A), the matched aesthetic prototype (the nearest one) and input face exhibit similar facial contours, offering effective reference for subsequent face generation. 

To show the effectiveness of the proposed NNFS module in matching faces by 3D facial structures, we compare the proposed NNFS module with Face++ \cite{MegEngine}.
Face++ is a deep learning-based framework for face searching.
Fig. \ref{Fig.7} presents the face beautification results of both methods. As shown in Fig. \ref{Fig.7}, Face++ is more sensitive to variations in facial expressions than NNFS. In contrast, the NNFS module, by decoupling facial identity features from facial expression features, is less affected by expressions. Consequently, the NNFS module tends to match more similar face structures than Face++. 
Quantitatively, NNFS achieves better ID preservation results compared to Face++ (refer to Table \ref{tab:table1_new}).
Due to the sensitivity to facial expression, pose, and lighting, the facial structure features matched by Face++ are often farther from the input image, thereby leading to excessive facial deformation in the generated images and lower ID Similarity.

\subsection{Effectiveness of Facial Guidance Extraction}
As introduced in Sec. \ref{sec:FGE}, the FGE module merges identity features of the input and reference faces with reconstructed 3D face model to obtain 3D structure guidance for FAE.
Significantly, the pose of the reference face is aligned with the input face in the reconstructed 3D domain and the projected 2D domain (i.e., view plane) to ensure the effectiveness of facial 3D structure guidance.
We further demonstrate the necessity of the FGE module through experiments, as depicted in Fig. \ref{fig:Module1_2_Comparative_experiment}(B). Qualitative results in Fig. \ref{fig:Module1_2_Comparative_experiment}(B) show that significant deviations in positions and directions of the generated face can be observed when the proposed FGE module is not employed.

Meanwhile, our research focuses on enhancing facial aesthetics through facial deformation. To mitigate the loss of identity caused by facial deformation, the proposed FGE module linearly blends the identity features of the input with reference faces, aiming to explicitly preserve identity (as introduced in Sec. \ref{sec:FGE}). To demonstrate its effectiveness, we compare the results of the 3D feature fusion method with those without the fusion method in the experiment. The qualitative comparison results are shown in Fig. \ref{Fig.8}, clearly demonstrating that facial deformation is more pronounced in the absence of feature fusion, leading to a decrease in ID Similarity. Hence, 3D feature fusion is significant to ensure the consistency of identity information. The quantitative comparison results in Table \ref{tab:table1_new} further support this conclusion. 

\begin{figure*}
    \centering
    \includegraphics[width=1\linewidth]{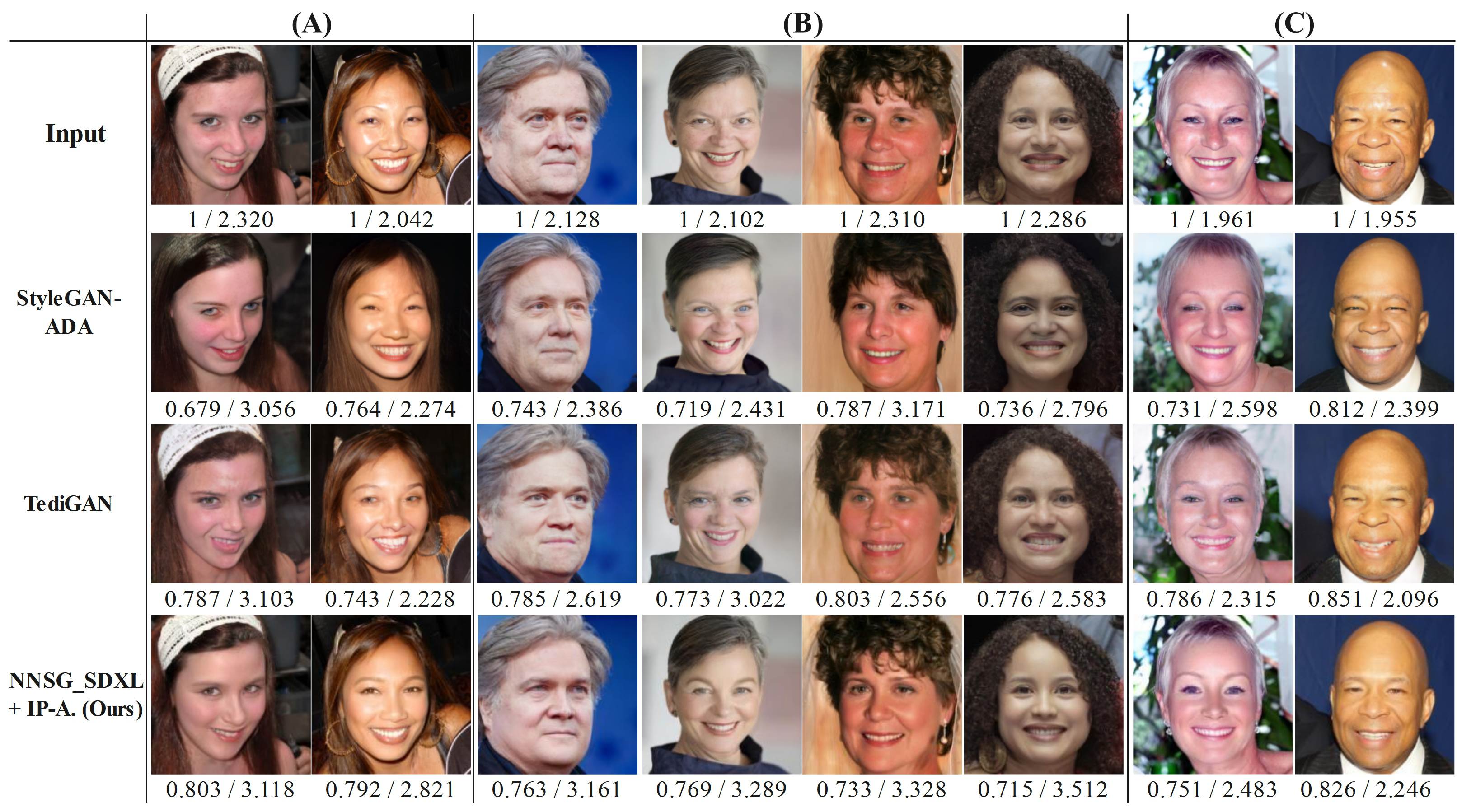}
    \caption{Qualitative comparison with GAN-based methods on FFHQ dataset. IP-A. stands for IP-Adapter, and the used version of Stable Diffusion model is SDXL-1.0. The scores below each image are ID Similarity / Attractiveness. We present the qualitative results in three typical groups: (A) the cases in which our method outperforms all contenders in both ID similarity and attractiveness; (B) the cases in which our method does not outperform all contenders in ID similarity; (C) the cases in which our method does not outperform all contenders in attractiveness.}
    \label{fig:peer_comparative_ffhq}
\end{figure*}

\subsection{Effectiveness of 3D Structure Guidance} 
\label{IV-D}
3D structure guidance involves two parts in the proposed method: the contour guidance representing the contours of the face on the view plane, which is obtained by Canny edge detector; and the depth guidance representing the face's depth information in the direction of the ray, which is extracted from the rendered 3D face.

Fig. \ref{fig:para_adject} reflects how the depth and the contour guidance contribute to the overall enhancement process respectively, where we gradually adjust the guidance weights from 0 to 1. 
In the first row of Fig. \ref{fig:para_adject}, as the control weight of contour guidance gradually increases, adjustments are made to the shape and position of the nose and eyes (as noted by blue boxes in the Fig. \ref{fig:para_adject} (a-c)), and the proportion of facial features (proportion among upper-facial, mid-facial and lower-facial regions as noted by dashed red lines) accordingly. The second row of Fig. \ref{fig:para_adject} demonstrates that depth guidance can be mainly utilized to control facial spatial structure, such as highlights on the forehead and shadow on the bridge of nose (as noted by red boxes in the Fig. \ref{fig:para_adject} (d-f)). Natural shadows and highlights are crucial for rendering the generated more realistic and lifelike facial structures. 

To demonstrate the effectiveness of the combination of different aspects of the proposed 3D structure guidance, we compare and analyze the attractiveness score and ID Similarity of face beautification with different guidance (see Fig. \ref{fig:exp_3d_structure}). When using SD for face generation without additional guidance, the random noise leads to a relatively high degree of randomness in the generated faces during the generation process. The high level of randomness results in faces that appear less realistic and natural. By combining both depth and contour guidance, the facial structure can be more precisely controlled, leading to higher ID similarity. Additionally, as the quantitative results are shown in Table \ref{tab:table1_new}, in terms of ID similarity, combining both depth and contour guidance outperforms the case when solely depth or contour guidance is used.

\begin{table}
\caption{Comparing different versions of Stable Diffusion model. The upward arrows indicate that larger values of the evaluation metric correspond to better results, while the downward arrows indicate the opposite. IP-A. stands for IP-Adapter.}
\begin{center}
\resizebox{1\columnwidth}{!}{
{\fontsize{46}{48}\selectfont
\begin{tabular}{llcccccc}
\toprule[4pt]
\multirow{2}{*}{Dataset}               & \multirow{2}{*}{Method} & \multicolumn{2}{c}{Attractiveness$\uparrow$} & \multirow{2}{*}{ID Sim.$\uparrow$} & \multirow{2}{*}{FID$\downarrow$} & \multirow{2}{*}{PSNR$\uparrow$} & \multirow{2}{*}{SSIM$\uparrow$} \\ \cmidrule(r){3-4}
                                       &                         & VGG16          & ResNet18          &                                 &                      &                       &                       \\ \midrule[2pt]
\multirow{3}{*}{FFHQ}                  & NNSG\_SD                & 3.055          & 2.659             & 0.749                           & 58.650               & \textbf{23.215}                & 0.753                 \\
                                       & NNSG\_SDXL             & \textbf{3.430}          & \textbf{2.976}             & 0.752                           & \textbf{24.739}               & 22.167                & 0.771                 \\ 
                                       & NNSG\_SDXL+IP-A.                       & 3.034          & 2.664   & \textbf{0.794}                                    & 26.535                 & 22.714                          & \textbf{0.801} \\ \midrule[0.5pt]
CelebAMask  & NNSG\_SD               & 2.912          & 2.560             & 0.696                           & 72.843               & \textbf{22.475}                & 0.732                 \\
-Hq (Mini)  & NNSG\_SDXL             & \textbf{3.360}          & \textbf{2.967}             & 0.703                           & 31.934               & 21.937                & 0.763                 \\  
            & NNSG\_SDXL+IP-A.                    & 3.171          & 2.688             & \textbf{0.806}                           & \textbf{30.543}               & 21.601                & \textbf{0.780} \\ \bottomrule[4pt]
\end{tabular}}}
\end{center}
\label{tab:table_SD_version}
\end{table}

\begin{table}
\caption{Peer comparison with GAN-based methods on FFHQ dataset. The upward arrows indicate that larger values of the evaluation metric correspond to better results, while the downward arrows indicate the opposite. IP-A. stands for IP-Adapter. The used version of Stable Diffusion model is SDXL-1.0.}
\begin{center}
\resizebox{1\columnwidth}{!}{
{\fontsize{26}{26}\selectfont
\begin{tabular}{lcccccc}
\toprule[2pt]
\multirow{2}{*}{Method} & \multicolumn{2}{c}{Attractiveness$\uparrow$} & \multirow{2}{*}{ID Sim.$\uparrow$} & \multirow{2}{*}{FID$\downarrow$} & \multirow{2}{*}{PSNR$\uparrow$} & \multirow{2}{*}{SSIM$\uparrow$} \\ \cmidrule(r){2-3}
                        & VGG16            & ResNet18        &                                 &                      &                       &                       \\ \midrule[1pt]
TediGAN \cite{xia2021open}              & 2.529      & 2.508           & \textbf{0.797}                  & 73.543               & 20.406                & 0.529                           \\
StyleGAN2-ADA \cite{karras2020training} & 2.533   & 2.460            & 0.782                                     & 95.978                           & 15.026                          & 0.557                           \\
NNSG\_SDXL (Ours)                       & \textbf{3.430}          & \textbf{2.976}   & 0.752                                     & \textbf{24.739}                  & 22.167                          & 0.771                           \\
NNSG\_SDXL+IP-A. (Ours)                       & 3.034          & 2.664   & 0.794                                    & 26.535                 & \textbf{22.714}                          & \textbf{0.801} \\ \bottomrule[2pt]
\end{tabular}}}
\end{center}
\label{tab:table_FFHQ}
\end{table}

\subsection{Sensitivity to Contour and Depth Guidance}
In Sec. \ref{IV-D}, we only studied the combination of depth or contour guidance by running the proposed FAE pipeline with or without them. To further study how generated results are sensitive to contour and depth guidance in a finer granularity, we quantitatively adjust the weights $(\eta, \omega)$ applied to contour and depth guidance introduced in Eq. (\ref{eq:guidance_fusion}). 

In the experiment, we set $\omega$ and $\eta$ ranging from 0 to 1, and analyze the variations of FID, Attractiveness, and ID Similarity. Three contour maps presented in Fig. \ref{fig:para_for_depth_contour} illustrate the influence of these two weight parameters. Fig. \ref{fig:para_for_depth_contour}(a) illustrates that a higher weight parameter of contour guidance leads to a higher FID value. The main reason for this phenomenon is that the reverse process of the diffusion model reduces noise by predicting the noise distribution of the forward process. Therefore, the extra guidance influences the noise distribution fitting in the reverse process and causes the increase of FID value. 

Meanwhile, the depth guidance enables precise control over facial spatial structure (highlights and shadows) and pose, making it particularly useful for generating images that preserve facial identity consistency (Fig. \ref{fig:para_for_depth_contour}(c)). Consequently, compared to the weight parameters of contour, the changes in the weight parameters of depth have a minor impact on the FID value (Fig. \ref{fig:para_for_depth_contour}(a)).

In terms of attractiveness, as shown in Fig. \ref{fig:para_for_depth_contour}(b), a larger weight value of the contour guidance leads enhanced face to be closer to the attractive reference face in terms of facial contours, leading to stronger beautification results. However, the generated results potentially lose the input's identity when the contour guidance is too strong, as shown in Fig. \ref{fig:para_for_depth_contour}(c). 

\begin{figure*}[t]
    \centering
    \includegraphics[width=1\textwidth]{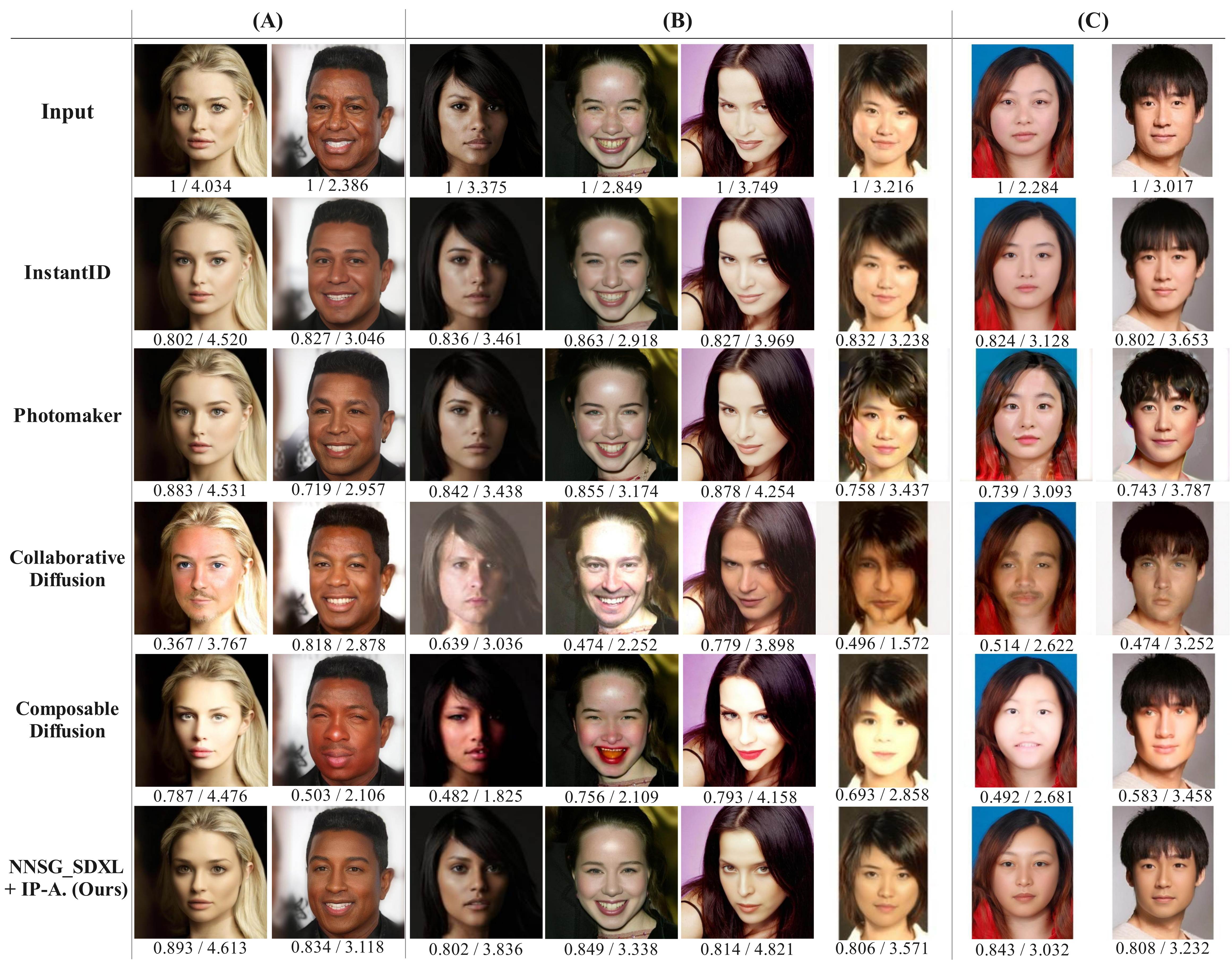}
    \caption{Qualitative comparison on CelebAMask-Hq (Large), CelebAMask-Hq (Mini), and SCUT-FBP datasets. The used version of Stable Diffusion model is SDXL-1.0 and IP-A. stands for IP-Adapter. The scores below each image are ID Similarity / Attractiveness. We present the qualitative results in three typical groups: (A) the cases in which our method outperforms all contenders in both ID similarity and attractiveness; (B) the cases in which our method does not outperform all contenders in ID similarity; (C) the cases in which our method does not outperform all contenders in attractiveness.}
    \label{fig:peer_comparative_CelebA_Large}
    \vspace{2mm}
\end{figure*}

\subsection{Use of different Stable Diffusion versions}
To compare the impacts of different SD versions on the effectiveness of the proposed framework, we conduct tests on two versions (SD-V1-2 and SDXL-1.0) separately, with an additional setting that combines SDXL-1.0 with IP-Adapter \cite{ye2023ipadapter}. We find that the SD-V1-2 version would cause over-represented red lips, skin tone and generate unnatural pupils. 
Compared to SD-V1-2, SDXL-1.0 or SDXL-1.0 with IP-Adapter  generates images with less aforementioned bias issues, as indicated by experimental results in Fig. \ref{fig:NNSG_Version_comparison} and Table \ref{tab:table_SD_version}.

\subsection{Comparative Study} \label{IV-F}
We present the study comparing our method to GAN-based approaches \cite{xia2021open, karras2020training} and diffusion-based approaches \mbox{\cite{huang2023collaborative, liu2022compositional, wang2024instantid, li2023photomaker}} for FAE as below:

\textbf{Contenders:} For a more comprehensive comparison with SOTA, we have employed six contender methods \mbox{\cite{xia2021open, karras2020training, huang2023collaborative, liu2022compositional, wang2024instantid, li2023photomaker}} for peer comparison. Two selected contender methods are based on GANs for facial attribute editing: TediGAN \cite{xia2021open} and StyleGAN2-ADA \cite{karras2020training}. These approaches blend input with reference facial styles to enhance the aesthetic appeal of input faces. 
The others, including Composable-Diffusion \cite{liu2022compositional}, Collaborative-Diffusion  \cite{huang2023collaborative}, InstantID \cite{wang2024instantid}, and Photomaker \cite{li2023photomaker}, are diffusion-based facial image editing techniques. 
These approaches transfer identity information from reference faces to input facial images via implicit visual features or explicit facial landmarks to guide the diffusion process. 

\textbf{Comparing to GAN-based approaches:} As shown in Fig. \ref{fig:peer_comparative_ffhq}, TediGAN \cite{xia2021open} and StyleGAN2-ADA \cite{karras2020training} employ blending operations on latent facial features, which lacks explicit control over fine-grained facial attributes. Specifically, StyleGAN2-ADA fails to maintain consistency in irrelevant attributes such as background and hairstyle across images. While TediGAN demonstrates some control over internal facial attributes, it remains insensitive to facial contours, limiting edits to coarse-grained features without sensitivity to fine-grained structures. In contrast, our method achieves more precise control over facial contours and spatial structure (shading and highlights), resulting in better aesthetic enhancement. 
Quantitative evaluation results in Table \ref{tab:table_FFHQ} indicate that images generated by our method receive higher attractiveness scores while maintaining identity consistency similar to TediGAN and better than StyleGAN2-ADA. 
To better analyze how the proposed method behaves differently from other methods, in Fig. \ref{fig:peer_comparative_ffhq}, we provide qualitative analysis in typical categories of cases: 1) the cases in which our method outperforms contenders in both ID similarity and attractiveness, as depicted in Fig. \ref{fig:peer_comparative_ffhq}(A); 2) the cases in which our method does not outperform all contenders in ID similarity, as depicted in Fig. \ref{fig:peer_comparative_ffhq}(B); 3) the cases in which our method does not outperform all contenders in attractiveness, as depicted in Fig. \ref{fig:peer_comparative_ffhq}(C).
In the first category (as Fig. \ref{fig:peer_comparative_ffhq}(A)), our method better improves facial attractiveness by smoothing facial texture, introducing shadows and highlights, and adjusting contours of eyes, noses, and lips. These adjustments are subtle and maintain facial identity. 
In the second category (as Fig. \ref{fig:peer_comparative_ffhq}(B)), when encountering inputs with obvious wrinkles on cheeks or around eyes, our method effectively removes these wrinkles, while this potentially lowers their ID similarities. 
In the third category (as Fig. \ref{fig:peer_comparative_ffhq}(C)), for some inputs which cannot be matched to an appropriate reference within our Aesthetic Prototype Database, the extracted guidance may not able to lead to sufficiently good results, but the attractiveness is improved comparing to the input.

\textbf{Comparing to diffusion-based methods:} The qualitative results comparing our method to diffusion-based contenders are given in Fig. \ref{fig:peer_comparative_CelebA_Large} and Table \ref{tab:table_peer_comparison}. 
InstantID introduces facial landmarks to the diffusion model as weak spatial constraints for facial adjustments. Because of the lack of fine-grained spatial guidance, such as the shadow of the nose bridge and external outline, these features from the reference face fail to transfer completely to the input face. 
Similarly, PhotoMaker is also insensitive to external contour features because of the lack of fine-grained spatial guidance. 
Conversely, our method focuses on structure control and employs facial deformation for enhancement, while some internal facial appearances of input images are potentially lost. 
Table \ref{tab:table_peer_comparison} shows that our method outperforms diffusion-based methods in attractiveness with ID Similarity results comparable to the contenders.
Similar to our analysis for GAN-based methods, we provide qualitative analysis in typical categories of cases in Fig. \ref{fig:peer_comparative_CelebA_Large} for diffusion-based methods.
Similarly, our method better improves facial attractiveness by smoothing facial texture, introducing shadows and highlights, and adjusting facial pose and contours of eyes, noses, and lips. These adjustments are subtle and maintain facial identity. While in the cases when encountering inputs with obvious wrinkles on cheeks or around eyes, our method effectively removes these wrinkles, while this potentially lowers their ID similarities. 
And in cases when the matched reference is not appropriate, the extracted guidance may not able to lead to the best results comparing to contenders, but the attractiveness is improved compared to the input.

\begin{figure}
    \centering
    \includegraphics[width=1\linewidth]{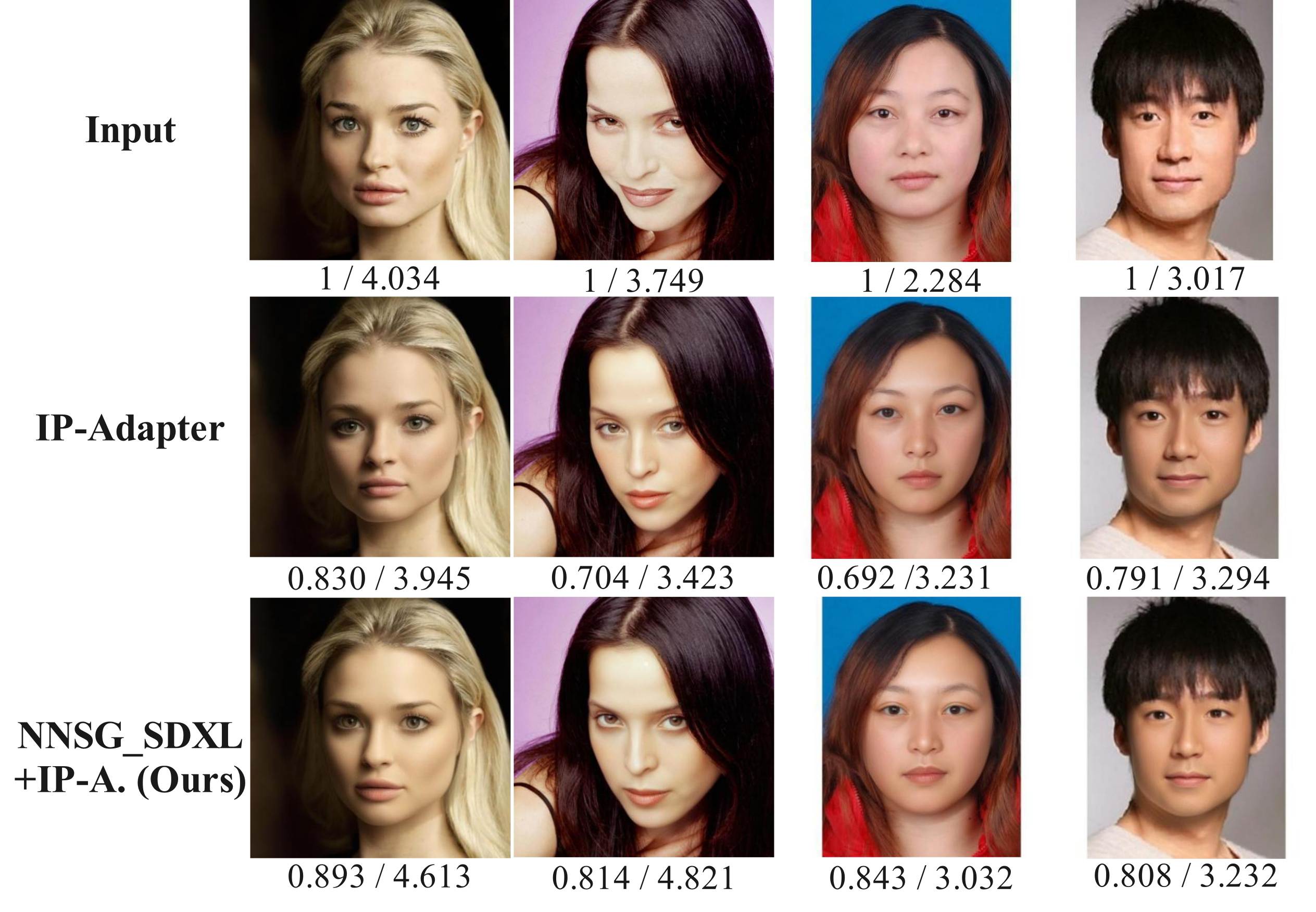}
    \caption{Qualitative analysis results comparing IP-Adapter alone (IP-Adapter) with the combining IP-Adapter with NNSG (NNSG\_SDXL+IP-A.) on CelebAMask-Hq (Large) and SCUT-FBP dataset. The used version of Stable Diffusion model is SDXL-1.0. The scores below each image are ID Similarity / Attractiveness.}
    \label{fig:IP-A_comparison}
\end{figure}

\textbf{Experimental results by diverse ages:} To show the effectiveness of our method across different age groups, we divide the test results on CelebAMask-Hq (Large) dataset into `old face' and `young face' categories according to the age labels provided by the CelebAMask-Hq dataset. The results in Table \ref{tab:table_age_and_race_comp} demonstrate that our method's performance shows more prominent decrease in ID similarity for `old face' images compared to `young face'. This is because our approach primarily uses facial depth and contours to guide the beautification process. As faces age, they usually develop wrinkles and experience muscle sagging, which are significant aesthetic features and also critical for facial identity. Our method, aimed at enhancing facial aesthetics, tends to remove such features to achieve a more aesthetically pleasing result. Consequently, this can lead to a loss of identity-related information and an excessive beautification effect in images of older faces. However, this may not be regarded as a drawback, since one of the objectives of facial aesthetics enhancement is to make people look younger.

\textbf{Experimental results by diverse ethnicities:} To demonstrate the effectiveness of our method across different ethnicities, we divide the test results into `Asian' and `Caucasian' categories according to the ethnicity labels in the SCUT-FBP dataset. Table \ref{tab:table_age_and_race_comp} shows that our method performs consistently well across these ethnic groups in terms of ID similarity and attractiveness. 
The attractiveness scores and facial identity similarities of the beautified faces remain relatively stable, indicating that our method effectively handles facial images from different ethnic backgrounds without significant variation in performance.

\begin{table}[t]
\caption{Peer comparison results with diffusion-based methods. The upward arrows indicate that larger values of the evaluation metric correspond to better results, while the downward arrows indicate the opposite. (Collab. Diff. stands for Collaborative Diffusion,  Comp. Diff. stands for Composable Diffusion, IP-A. stands for IP-Adapter. The used version of Stable Diffusion model is SDXL-1.0.)}
\begin{center}
\resizebox{1\columnwidth}{!}{
{\fontsize{46}{48}\selectfont
\begin{tabular}{llcccccc}
\toprule[4pt]
\multirow{2}{*}{Dataset}               & \multirow{2}{*}{Method} & \multicolumn{2}{c}{Attractiveness$\uparrow$} & \multirow{2}{*}{ID Sim.$\uparrow$} & \multirow{2}{*}{FID$\downarrow$} & \multirow{2}{*}{PSNR$\uparrow$} & \multirow{2}{*}{SSIM$\uparrow$} \\ \cmidrule(r){3-4}
                                       &                         & VGG16          & ResNet18          &                                 &                      &                       &                       \\ \midrule[2pt]
                                       & Collab. Diff. \cite{huang2023collaborative} & 3.540          & 3.282             & 0.680                           & 31.070               & 16.532                & 0.640                 \\
                                       & Comp. Diff. \cite{liu2022compositional}     & 3.802          & 3.427             & 0.694                           & 23.405               & 21.146                & 0.744                 \\
CelebAMask                             & InstantID \cite{wang2024instantid}          & 3.706          & 3.429             & 0.827                           & \textbf{14.540}      & 25.290                & 0.743                 \\
-Hq(Large)                             & PhotoMaker \cite{li2023photomaker}          & 3.856          & 3.501             & \textbf{0.849}                  & 18.147               & \textbf{26.205}       & 0.792                 \\
                                       & NNSG\_SDXL (Ours)                           & 3.884          & 3.507             & 0.747                           & 19.450               & 22.803                & 0.805                 \\ 
                                       & NNSG\_SDXL+IP-A. (Ours)                     & \textbf{3.898} & \textbf{3.560}    & 0.812                           & 21.494               & 22.773                & \textbf{0.824}        \\ \midrule[2pt]
\multirow{6}{*}{SCUT-FBP}              & Collab. Diff.\cite{huang2023collaborative}  & 3.293          & 2.969             & 0.637                           & 34.794               & 16.062                & 0.726                 \\
                                       & Comp. Diff.\cite{liu2022compositional}      & 3.513          & 3.113             & 0.622                           & 20.134               & 21.108                & 0.862                 \\
                                       & InstantID \cite{wang2024instantid}          & 3.419          & 3.161             & 0.836                           & \textbf{8.722}       & \textbf{28.207}       & 0.876                 \\
                                       & PhotoMaker \cite{li2023photomaker}          & 3.405          & 3.202             & 0.782                           & 23.955               & 23.789                & 0.791                 \\
                                       & NNSG\_SDXL(Ours)                            & \textbf{3.539}  & \textbf{3.340}   & 0.756                           & 15.236               & 23.686                & 0.883                 \\ 
                                       & NNSG\_SDXL+IP-A. (Ours)                     & 3.423          & 3.218             & \textbf{0.852}                  & 19.310               & 24.557                & \textbf{0.903}        \\ \midrule[2pt]
                                       & Collab. Diff.\cite{huang2023collaborative}  & 2.700          & 2.711             & 0.644                           & 51.012               & 15.867                & 0.601                 \\
                                       & Comp. Diff.\cite{liu2022compositional}      & 2.751          & 2.639             & 0.568                           & 47.057               & 22.024                & 0.716                 \\
CelebAMask                             & InstantID \cite{wang2024instantid}          & 2.809          & 2.628             & \textbf{0.851}                  & \textbf{24.562}      & \textbf{24.876}       & 0.718                 \\
-Hq(Mini)                              & PhotoMaker \cite{li2023photomaker}          & 2.896          & 2.624             & 0.816                           & 26.731               & 23.217                & 0.675                 \\
                                       & NNSG\_SDXL(Ours)                            & \textbf{3.360} & \textbf{2.967}    & 0.703                           & 31.934               & 21.937                & 0.763 \\ 
                                       & NNSG\_SDXL+IP-A. (Ours)                     & 3.171          & 2.688             & 0.806                           & 30.543               & 21.601                & \textbf{0.780} \\ \bottomrule[4pt]
\end{tabular}}}
\end{center}
\label{tab:table_peer_comparison}
\end{table}

\subsection{Combination with implicit ID-preservation methods}

Our method is designed to mainly focus on manipulating 3D facial structures (e.g., contours and depth).
However, it is undeniable that facial appearances (e.g.,  skin tone and lip color) and 3D facial structures play equally important roles in identity preservation and facial aesthetics enhancement.

IP-Adapter \cite{ye2023ipadapter} and similar identity preservation methods primarily concentrate on generating internal facial appearances and may not be sensitive to 3D facial structure. 
Hence, our approach can be integrated with ID-preservation methods focusing on inner facial appearances such as IP-Adapter \cite{ye2023ipadapter} to provide effective FAE guidance in both 3D structure and 2D appearance. 
As shown in row three of Fig. \ref{fig:IP-A_comparison}, the combining method (NNSG\_SDXL+IP-Adapter) can generate better results. 
Table \ref{tab:table_SDXL_IP-Adapter_basel} and Fig. \ref{fig:IP-A_comparison} show the comparison results of IP-Adapter alone and the combination method (NNSG\_SDXL+IP-Adapter). 
These results demonstrate that the combined method (NNSG\_SDXL+IP-Adapter) can effectively improve generated facial ID similarity while achieving higher aesthetic enhancement performance.

\label{sec:sub_exp_results}
\subsection{Subjective Experiment by User Surveys}

To further confirm the aesthetic improvements brought by our method, we conduct an additional subjective evaluation. 
We select 100 facial images from the CelebAMask-Hq and SCUT-FBP datasets, ensuring a diverse data selection with 25 random images each for old faces, young faces, Asian faces, and Caucasian faces, respectively. For the subjective evaluation, these images were enhanced using our method and four other contenders: our NNSG-Diffusion (with SDXL and IP-Adapter), InstantID, Photomaker, Collaborative Diffusion, and Composable Diffusion. These five methods produced five beautified images per original image, totaling 100 sets of images (25 per category). 

\begin{table}[t]
\caption{Peer comparison results across different age and ethnicity categories. The upward arrows indicate that larger values of the evaluation metric correspond to better results, while the downward arrows indicate the opposite. (Collab. Diff. stands for Collaborative Diffusion,  Comp. Diff. stands for Composable Diffusion, IP-A. stands for IP-Adapter. The used version of Stable Diffusion model is SDXL-1.0.)}
\begin{center}
\resizebox{1\columnwidth}{!}{
{\fontsize{46}{46}\selectfont
\begin{tabular}{llllcccccc}
\toprule[5pt]
\multirow{2}{*}{Dataset}               & \multirow{2}{*}{Method} & \multicolumn{2}{c}{Attractiveness$\uparrow$} & \multirow{2}{*}{ID Sim.$\uparrow$} & \multirow{2}{*}{FID$\downarrow$} & \multirow{2}{*}{PSNR$\uparrow$} & \multirow{2}{*}{SSIM$\uparrow$} \\ \cmidrule(r){3-4}
                                       &                         & VGG16          & ResNet18          &                                 &                      &                       &                       \\ \midrule[0.5pt]
\multicolumn{1}{c}{}                           & Collab. Diff.\cite{huang2023collaborative}&  3.397 &  3.148 &  0.652    &  16.948 &  16.809 &  0.636 \\
\multicolumn{1}{c}{}                           & Comp. Diff.\cite{liu2022compositional}    &  3.654 &  3.273 &  0.647    &  24.669 &  20.931 &  0.734 \\
\multicolumn{1}{c}{}                           & InstantID  \cite{wang2024instantid}       &  3.523 &  3.256 &  0.818    &  \textbf{14.545} &  25.250 &  0.736 \\
\multicolumn{1}{c}{}                           & Photomaker \cite{li2023photomaker}        &  3.670 &  3.317 &  \textbf{0.851}    &  25.362 &  \textbf{26.228} &  0.792 \\
\multicolumn{1}{c}{}                           & NNSG\_SDXL(Ours)                          &  \textbf{3.759} &  3.381    &  0.714    &  22.923 &  22.956 &  \textbf{0.816} \\ 
\multicolumn{1}{c}{\multirow{-6}{*}{Old face}} & NNSG\_SDXL+IP-A.(Ours)                    &  3.740          &  \textbf{3.423} &  0.747    &  23.579 &  22.445 & 0.807 \\ \midrule[2.5pt]
                                               & Collab. Diff.\cite{huang2023collaborative}&  3.641 &  3.376 &  0.703    &  16.345 &  16.339 &  0.642 \\
                                               & Comp. Diff.\cite{liu2022compositional}    &  3.906 &  3.535 &  0.739    &  23.464 &  21.296 &  0.751 \\
                                               & InstantID  \cite{wang2024instantid}       &  3.836 &  3.552 &  0.833    &  \textbf{14.335} &  25.318 &  0.749 \\
                                               & Photomaker  \cite{li2023photomaker}       &  3.960 &  3.550 &  \textbf{0.846}    &  25.657 &  \textbf{26.188} &  0.792 \\
                                               & NNSG\_SDXL(Ours)                          &  \textbf{3.975} &  3.599 &  0.779    &  22.038 &  22.592 &  0.789 \\ 
\multirow{-6}{*}{Young face}                   & NNSG\_SDXL+IP-A.(Ours)                    &  3.974 & \textbf{3.626} &  0.796    &  24.442 &  22.932 &  \textbf{0.832} \\ \midrule[2.5pt]
\multicolumn{1}{c}{}                           & Collab. Diff.\cite{huang2023collaborative}   &   3.399 &   3.054 &   0.635    &   31.882 &   16.441 &   0.738 \\
\multicolumn{1}{c}{}                           & Comp. Diff.\cite{liu2022compositional}       &   3.509 &   3.104 &   0.619    &   23.985 &   21.628 &   0.871 \\
\multicolumn{1}{c}{}                           & InstantID  \cite{wang2024instantid}          &   3.428 &   3.170 &   0.850    &   \textbf{9.712}  &   \textbf{28.346} &   0.879 \\
\multicolumn{1}{c}{}                           & Photomaker \cite{li2023photomaker}           &   3.469 &   3.226 &   0.779    &   23.943 &   23.822 &   0.792 \\
\multicolumn{1}{c}{}                           & NNSG\_SDXL(Ours)                             &   \textbf{3.523} &   \textbf{3.339} &  0.757    &   15.947 &   23.849 &  0.888 \\ 
\multicolumn{1}{c}{\multirow{-6}{*}{Asian}}    & NNSG\_SDXL+IP-A.(Ours)                       &   3.356 & 3.152 &  \textbf{0.854}    &  18.875  &   24.586 &  \textbf{0.904} \\ \midrule[2.5pt]
                                               & Collab. Diff.\cite{huang2023collaborative}  &   3.011 &   2.744 &   0.642    &   31.042 &   15.052 &   0.698 \\
                                               & Comp. Diff.\cite{liu2022compositional}      &   3.525 &   3.137 &   0.630    &   23.180 &   19.723 &   0.837 \\
                                               & InstantID  \cite{wang2024instantid}         &   3.394 &   3.168 &   0.800    &   \textbf{8.146}  &   \textbf{27.834} &   0.866 \\
                                               & Photomaker  \cite{li2023photomaker}         &   3.343 &   3.138 &   0.791    &   23.246 &   23.699 &   0.789 \\
                                               & NNSG\_SDXL(Ours)                            &   3.629 &   3.347 &   0.750    &   15.138 &   22.754 &   0.854 \\ 
\multirow{-6}{*}{Caucasian}                    & NNSG\_SDXL+IP-A.(Ours)                      &   \textbf{4.154} &   \textbf{3.945} &   \textbf{0.828}    &   17.532 &   24.236 &   \textbf{0.888} \\ \bottomrule[5pt]
\end{tabular}}}
\end{center}
\label{tab:table_age_and_race_comp}
\end{table}

\begin{figure}
    \centering
    \includegraphics[width=1\linewidth]{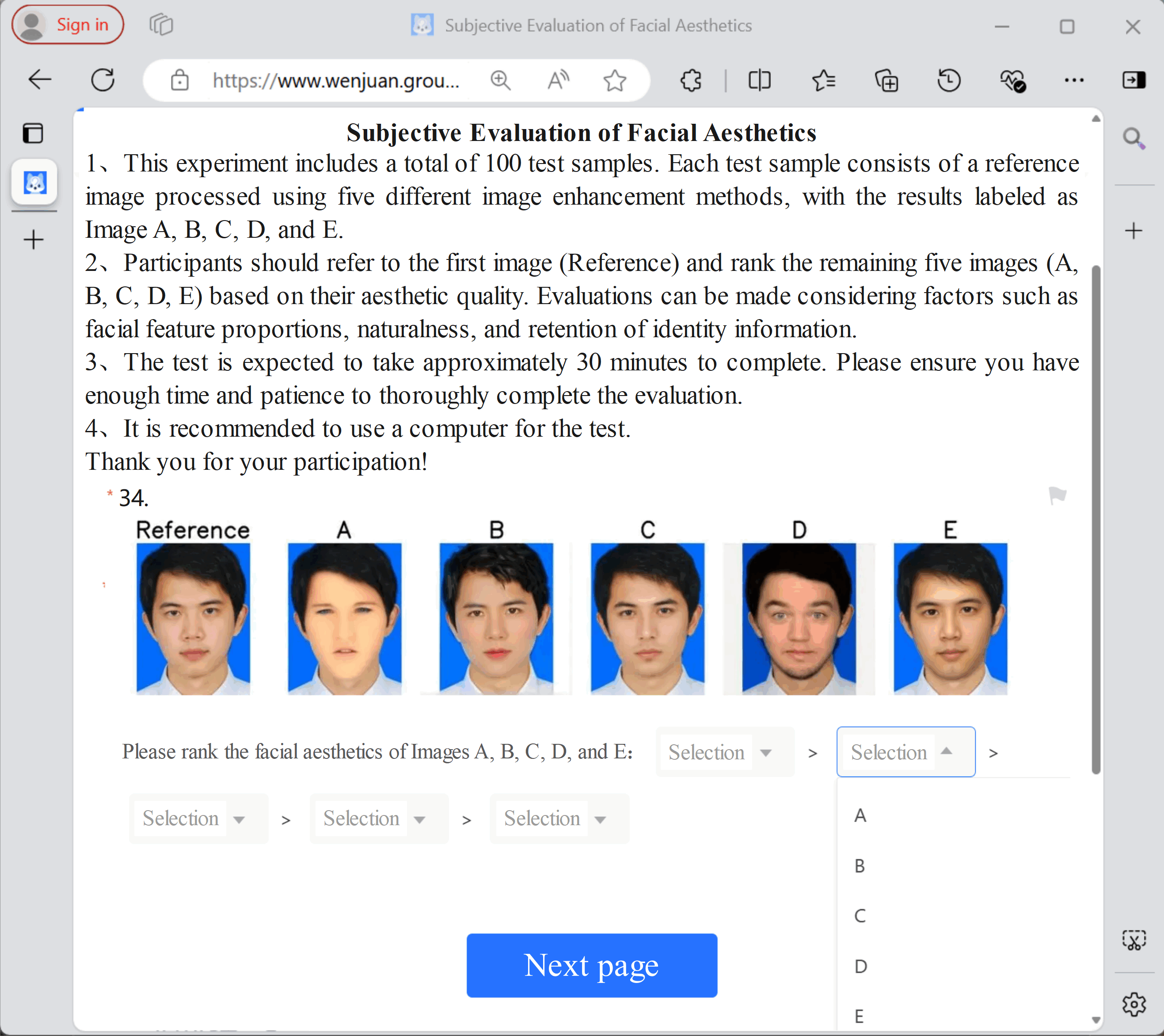}
    \caption{Demonstration of UI for our user survey, where the original UI is in Chinese. (Methods for results from A to E: Composable Diffusion, Photomaker, InstantID, Collaborative Diffusion, NNSG\_SDXL+IP-Adapter(Ours)).}
    \label{fig:sur_screen}
\end{figure}

We engaged 20 participants (8 females, 12 males) to rank the attractiveness of the images in each set from highest to lowest based on facial aesthetics. 
The participants' ages range from 15 to 50 years. They come from a variety of professional backgrounds, are all Chinese, and have normal vision. 
In our survey, we randomly shuffled the order of the five enhancement results as options for each question, using the original image as a reference. 
In the instructions, we ask participants to rank the given face images based on facial attractiveness. Specifically, participants are asked to evaluate facial attractiveness by considering facial aesthetics, naturalness, and identity consistency compared to the original image references.
A demonstration of the survey user interface is shown in Fig. \ref{fig:sur_screen}. The ranking was scored with 5 points for the most attractive, down to 1 point for the least attractive (step size is one).

The average scores for each method on each category of data are presented in Table \ref{tab:table_sub_exp}. 
Our method achieves the highest subjective score among the five methods, and outperforms Collaborative Diffusion and Composable Diffusion by a large margin, demonstrating its superior performance in enhancing facial aesthetics.

\section{Discussions about Limitations}
\textbf{Lighting Variations:} Our method utilizes depth guidance to control shadows and highlights in the generated facial images. This approach affects the diffusion model's rendering of lighting, leading to variations in the lighting of the generated faces. Although changes in lighting can influence the naturalness of the images, small-scale subjective experiments (in Table \ref{tab:table_sub_exp}) have revealed that these variations have less impact on the users' overall attractiveness evaluation on the results of our method. 

{\textbf{Focus on 3D Facial Structure:} The visual results of Fig. \ref{fig:NNSG_Version_comparison} (row 2 and row 3) occasionally show inconsistencies in identity preservation and an overemphasis on beautification subjectively when only the proposed NNSG-Diffusion is adopted. 
This shows one limitation that our approach primarily affects the contours, pose, and spatial structure of the input face to enhance facial aesthetics while not considering variations in internal facial appearances (e.g. skin tone, and colour of eyes). 
However, internal facial appearances are also important for preserving individual identity. Hence, there is a potential risk when neglecting consistency maintenance in colour of facial appearance, which impacts subjective assessment for identity consistency. 
By incorporating our method with IP-Adapter, these internal facial appearances can be considered during the enhancement process for better identity preservation (as shown in Table \ref{tab:table_SDXL_IP-Adapter_basel}, Table \ref{tab:table_SD_version} and Fig. \ref{fig:IP-A_comparison}). 
Meanwhile, IP-Adapter \cite{ye2023ipadapter} and similar identity preservation methods primarily concentrate on generating internal facial appearances and may not be sensitive to facial structure features. Our method, combined with an IP-Adapter, can effectively relieve this problem and minimize the loss of facial contour information (as shown in Fig. \ref{fig:IP-A_comparison} and Table \ref{tab:table_SDXL_IP-Adapter_basel}).}

\textbf{Limited reference faces}: Currently, the limited size of our aesthetic prototype dataset may lead to cases where the matched aesthetic prototype is not sufficiently close to the input face. 
The enhancement result may become less similar to the input face in terms of identity
when using such aesthetic prototype as reference, and the enhancement result may become less similar to the input face in terms of identity.
To address this issue, a larger dataset can provide a broader range of aesthetic prototypes for future usage, allowing for a better match between the reference face images and the input face. This improvement would help reduce the impact on identity preservation brought by the enhancement process, ensuring that the enhanced face maintains a higher degree of consistency with the original identity while still achieving the desired aesthetic enhancement.

\section{Conclusion}

In this work, we aim for a facial aesthetics enhancement (FAE) method that provides adjustments to facial structures and preserves the identity of the input face as much as possible.

\begin{table}
\caption{Results comparing IP-Adapter alone (w/o NNSG) with the combining IP-Adapter with NNSG (w/ NNSG) on SCUT-FBP dataset. The upward arrows indicate that larger values of the evaluation metric correspond to better results, while the downward arrows indicate the opposite. The used version of Stable Diffusion model is SDXL-1.0.}
\begin{center}
\resizebox{1\columnwidth}{!}{
{\fontsize{16}{14}\selectfont
\begin{tabular}{lcccccc}
\toprule[1pt]
\multirow{2}{*}{Method} & \multicolumn{2}{c}{Attractiveness$\uparrow$} & \multirow{2}{*}{ID Sim.$\uparrow$} & \multirow{2}{*}{FID$\downarrow$} & \multirow{2}{*}{PSNR$\uparrow$} & \multirow{2}{*}{SSIM$\uparrow$} \\ \cmidrule(r){2-3}
                        & VGG16            & ResNet18        &                                 &                      &                       &                       \\ \midrule[0.5pt]
w/o NNSG                & 3.380            & 3.129           & 0.840                           & \textbf{16.160}               & 22.534                & 0.878                 \\
w/ NNSG                 & \textbf{3.423}   & \textbf{3.218}  & \textbf{0.852}                  & 19.310      & \textbf{24.557}       & \textbf{0.903}        \\ \bottomrule[1pt]
\end{tabular}
}}
\end{center}
\label{tab:table_SDXL_IP-Adapter_basel}
\end{table}

\begin{table}[t]
\caption{Average ratings of subjective evaluation results across four categories. (Collab. Diff. stands for Collaborative Diffusion,  Comp. Diff. stands for Composable Diffusion, IP-A. stands for IP-Adapter. The used version of Stable Diffusion model is SDXL-1.0.)}
\begin{center}
\resizebox{1\columnwidth}{!}{
{\fontsize{28}{14}\selectfont
\begin{tabular}{lccccc}
\toprule[2pt]
\multirow{2}{*}{Dataset}   & NNSG\_SDXL    & \multirow{2}{*}{InstantID}     & \multirow{2}{*}{Photomaker}    & \multirow{2}{*}{Collab. Diff.}   & \multirow{2}{*}{Comp. Diff.} \\ 
   & +IP-A.(Ours)    &      &     &    &  \\ \midrule[0.5pt]
Old       & \textbf{3.804}   & 3.735         & 3.700         & 1.880                     & 1.881                \\
Young     & \textbf{3.887}   & 3.722         & 3.808         & 1.597                     & 1.987                \\
Asian     & \textbf{3.934}   & 3.819         & 3.704         & 1.551                     & 1.992                \\
Caucasian & \textbf{3.847}   & 3.717         & 3.828         & 1.451                     & 2.157                \\
Total     & \textbf{3.868}   & 3.748         & 3.760         & 1.620                     & 2.004                \\ \bottomrule[2pt] 
\end{tabular}
}}
\end{center}
\label{tab:table_sub_exp}
\end{table}
 
\noindent Specifically, by extracting 3D structure guidance from a nearest neighbor reference face within an aesthetic prototype database we have established, our method allows the downstream FAE process to introduce fewer changes to the input's identity but ensures improvements in facial aesthetics.
The 3D structure guidance we have extracted consists of both depth and contour clues from 3D face models reconstructed from the reference face and the input face, which provides effective guidance to Stable Diffusion with ControlNet to conduct the facial beautification process.
Ablation studies demonstrated that our method effectively brings aesthetic improvements with less identity loss.
Peer comparisons were conducted with both GAN-based and diffusion-based methods. In contrast to these approaches, finer control over facial structure is achieved by our method through 3D structure guidance. This capability includes adjustments to facial pose, improvements in facial feature proportions, and precise deformation of facial contours. 
To further substantiate the aesthetic improvements brought by our method, we conduct an additional subjective evaluation. Our method achieves the highest subjective scores among the diffusion-based methods, demonstrating its performance in enhancing facial aesthetics.

\bibliographystyle{IEEEtran}
\bibliography{Manuscript.bbl}

\vspace{130 mm} 
\begin{IEEEbiography}
    [{\includegraphics[width=1in,height=1.25in, clip,keepaspectratio]{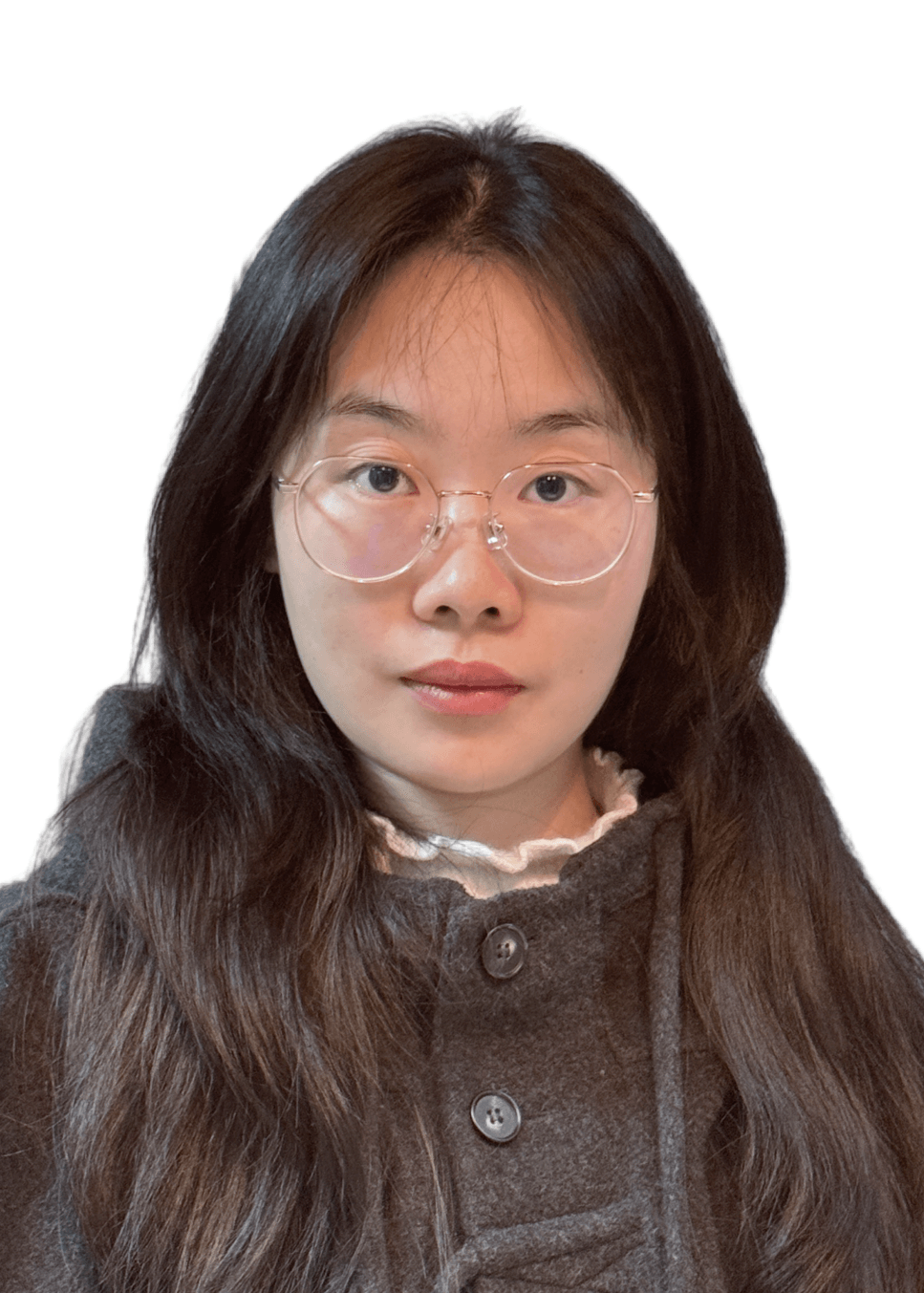}}]{Lisha Li} received her bachelor’s and master's degree from East China Jiaotong University, Nanchang, China. She is currently pursuing the Ph.D. degree with the School of Computing and Artificial Intelligence, Jiangxi University of Finance and Economics, Nanchang, China. Her research interests include visual quality assessment and computer vision.
\end{IEEEbiography}
\vspace{-10 mm} 

\begin{IEEEbiography}
    [{\includegraphics[width=1in,height=1.25in, clip,keepaspectratio]{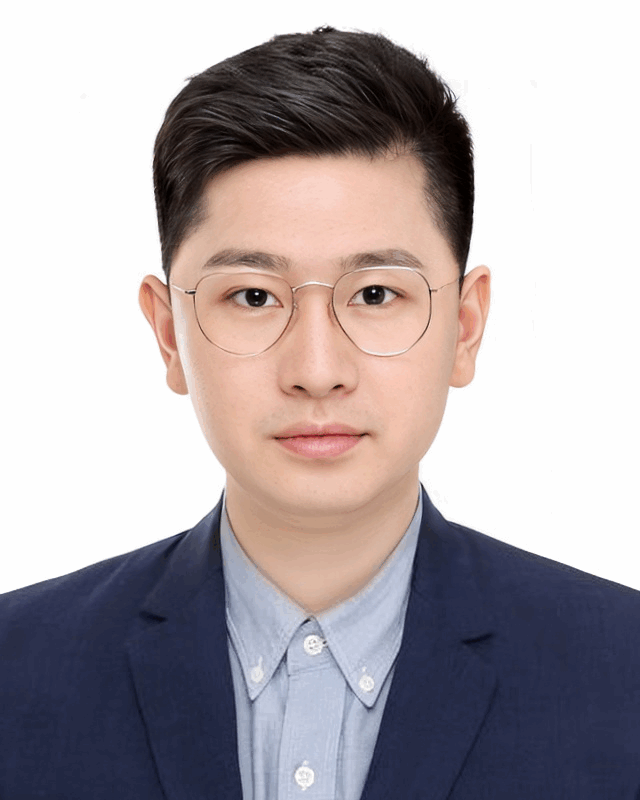}}]{Jingwen Hou} received his Ph.D. from Nanyang Technological University, Singapore. Prior to that, he obtained his Bachelor’s degree from the Beijing University of Posts and Telecommunications, Beijing, China, and his Master’s degree from Carnegie Mellon University, Pittsburgh, USA, both in E-Business. He is currently with the School of Computing and Artificial Intelligence, Jiangxi University of Finance and Economics, Nanchang, China.  His research interests include multimedia quality assessment and computer vision.
\end{IEEEbiography}
\vspace{-10 mm} 

\begin{IEEEbiography}
    [{\includegraphics[width=1in,height=1.25in, clip,keepaspectratio]{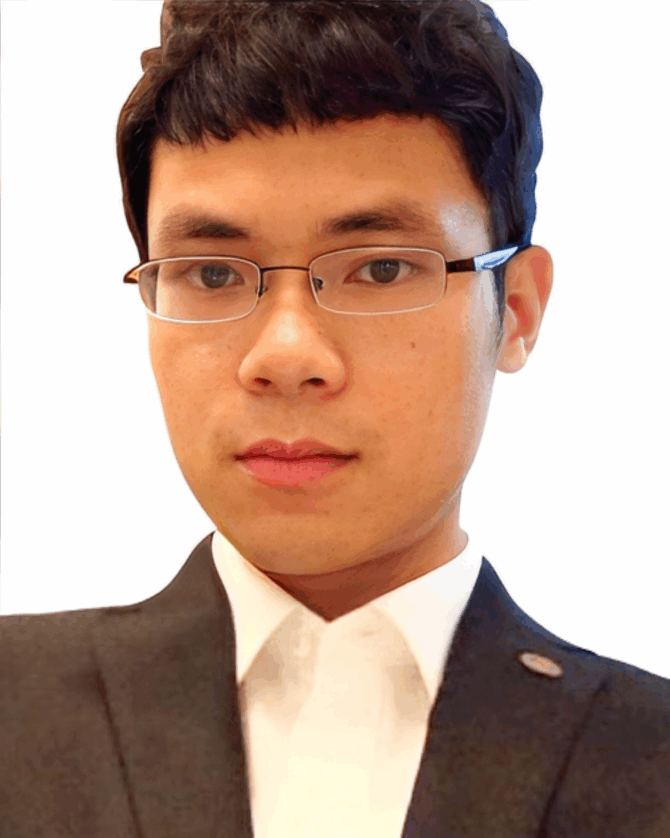}}]{Weide Liu} serves as a Research Fellow at Boston Children's Hospital, Harvard Medical School. Previously, he held the position of Research Scientist at A*STAR in Singapore. Dr. Liu earned both his Ph.D. and Bachelor's degrees from Nanyang Technological University. His research primarily focuses on computer vision, natural language processing, machine learning, and medical image analysis.
\end{IEEEbiography}
\vspace{-10 mm} 

\begin{IEEEbiography}
    [{\includegraphics[width=1in,height=1.25in, clip,keepaspectratio]{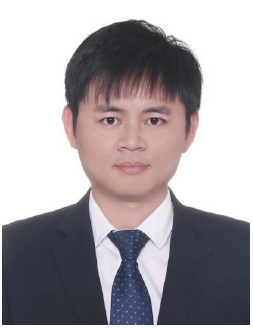}}]{Yuming Fang} (Senior Member, IEEE) received the B.E. degree from Sichuan University, Chengdu, China, the M.S. degree from the Beijing University of Technology, Beijing, China, and the Ph.D. degree from Nanyang Technological University, Singapore. He is currently a Professor with the School of Computing and Artificial Intelligence, Jiangxi University of Finance and Economics, Nanchang, China. His research interests include visual attention modeling, visual quality assessment, computer vision, and 3D image/video processing. He serves on the editorial board for \textsc{IEEE Transactions on Multimedia} and \textsc{Signal Processing: Image Communication}
\end{IEEEbiography}
\vspace{-10 mm} 

\begin{IEEEbiography}
    [{\includegraphics[width=1in,height=1.4in, clip]{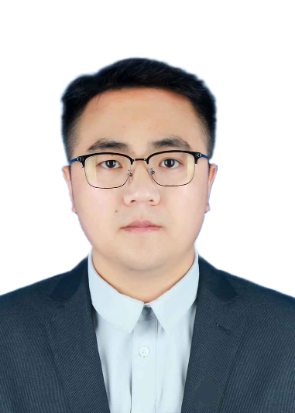}}]{Jiebin Yan} received the Ph.D. degree from Jiangxi University of Finance and Economics, Nanchang, China. He was a computer vision engineer with MTlab, Meitu. Inc, and a research intern with MOKU Laboratory, Alibaba Group. From 2021 to 2022, he was a visiting Ph.D. student with the Department of Electrical and Computer Engineering, University of Waterloo, Canada. He is currently a Lecturer with the School of Computing and Artificial Intelligence, Jiangxi University of Finance and Economics, Nanchang, China. His research interests include visual quality assessment and computer vision.
\end{IEEEbiography}

\end{document}